\newif\ifisTR
\isTRtrue

\documentclass{article} % For LaTeX2e
\usepackage{fullpage}

\usepackage[utf8]{inputenc} % allow utf-8 input
\usepackage[T1]{fontenc}    % use 8-bit T1 fonts
\usepackage{url}            % simple URL typesetting
\usepackage{booktabs}       % professional-quality tables
\usepackage{amsfonts}       % blackboard math symbols
\usepackage{nicefrac}       % compact symbols for 1/2, etc.
\usepackage{microtype}      % microtypography

\usepackage{subcaption}
\usepackage[dvipsnames]{xcolor}
\usepackage{color, colortbl}
\usepackage{multirow}
\usepackage{enumitem}
\usepackage{placeins}
\usepackage{graphicx}
\usepackage{booktabs} % for professional tables
\definecolor{LightBlue}{HTML}{F3F3F3}
\definecolor{dullGray}{HTML}{F3F3F3}
\definecolor{smallColorT}{HTML}{fff1e6}
\definecolor{largeColorT}{HTML}{edf2fb}
\definecolor{composeColorT}{HTML}{fff5ff}

\usepackage{utfsym}
\usepackage{amsmath}
\usepackage{amssymb}
\usepackage{mathtools}
\usepackage{amsthm}
\usepackage{xspace}
\usepackage{minitoc}
\usepackage{subcaption}
\usepackage{wrapfig}
\usepackage{overpic}
\usepackage{algpseudocode}
\usepackage[round,semicolon,sort]{natbib}
\usepackage[colorlinks,citecolor=mydarkblue,urlcolor=mydarkblue,linkcolor=mydarkblue]{hyperref}
\usepackage{tcolorbox}
\usepackage{microtype}
\usepackage{nicefrac} 
\usepackage[capitalize,noabbrev]{cleveref}

\theoremstyle{plain}

\theoremstyle{definition}

\theoremstyle{remark}

\newcommand{\ALPHA}{\texttt{PL\_Alpha}\xspace}
\newcommand{\ALPHAHILL}{\texttt{PL\_Alpha\_Hill}\xspace}
\newcommand{\ALPHAHAT}{\texttt{Alpha\_Hat}\xspace}

\newcommand{\NORM}{\texttt{Frobenius\_Norm}\xspace}

\newcommand{\LOGSPECTRALNORM}{\texttt{log\_spectral\_norm}\xspace}
\newcommand{\SPECTRALNORM}{\texttt{Spectral\_Norm}\xspace}

\newcommand{\STABLERANK}{\texttt{Stable\_Rank}\xspace}

\newcommand{\ENTROPY}{\texttt{Entropy}\xspace}

\renewcommand{\baselinestretch}{0.98}

\definecolor{mydarkblue}{rgb}{0,0.08,0.45}

\renewcommand{\cite}[1]{\citep{#1}}
\newcommand{\ourmethod}{\texttt{AlphaPruning}\xspace}

\begin{document}

\title{AlphaPruning: Using Heavy-Tailed Self Regularization Theory for Improved Layer-wise Pruning of Large Language Models}
\date{}
\author{
  Haiquan Lu$^{1}\footnote{First two authors contributed equally.}$, 
  Yefan Zhou$^{2}\footnotemark[1]$, 
  Shiwei Liu$^{3}$,  \\
  Zhangyang Wang$^{4}$, 
  Michael W. Mahoney$^{5, 6, 7}$, 
  Yaoqing Yang$^{2}$ \\
  $^1$ Nankai University\\
  $^2$ Dartmouth College\\
  $^3$ University of Oxford\\
  $^4$ University of Texas at Austin\\
  $^5$ International Computer Science Institute \\
  $^6$ Lawrence Berkeley National Laboratory \\
  $^7$ University of California at Berkeley \\
}

\maketitle

\begin{abstract}
    Recent work on pruning large language models (LLMs) has shown that one can eliminate a large number of parameters without compromising performance, making pruning a promising strategy to reduce LLM model size.
    Existing LLM pruning strategies typically assign uniform pruning ratios across layers, limiting overall pruning ability; and recent work on layerwise pruning of LLMs is often based on heuristics that can easily lead to suboptimal performance.
    In this paper, we leverage Heavy-Tailed Self-Regularization (HT-SR) Theory, in particular the \emph{shape} of empirical spectral densities (ESDs) of weight matrices, to design improved layerwise pruning ratios for LLMs.
    Our analysis reveals a wide variability in how well-trained, and thus relatedly how prunable, different layers of an LLM are.
    Based on this, we propose \ourmethod, which uses shape metrics to allocate layerwise sparsity ratios in a more theoretically-principled manner.
    \ourmethod can be used in conjunction with multiple existing LLM pruning methods.   
    Our empirical results show that \ourmethod prunes LLaMA-7B to 80\% sparsity while maintaining reasonable perplexity, marking a first in the literature on LLMs. 
    We have open-sourced our code.\footnote{\href{https://github.com/haiquanlu/AlphaPruning}{https://github.com/haiquanlu/AlphaPruning}}
\end{abstract}

\section{Introduction}\label{sec:intro}

Recent work on pruning large language models (LLMs)~\cite{jaiswal2023emergence, frantar2023massive, sun2023simple} has shown the ability to reduce the number of parameters significantly, without compromising performance, resulting in notable savings in memory footprint, computing time, and energy consumption.
Unlike pre-LLM pruning methods~\cite{sanh2020movement, kurtic2022optimal}, existing LLM pruning approaches typically allocate the ``sparsity budget'' (i.e., the number of pruned parameters or pruning ratios) uniformly across layers, making it difficult to increase sparsity to very high levels.
Relatively little effort has been put into developing theoretically-principled ways to compute layerwise pruning ratios.
For example, the Outlier Weighed Layerwise sparsity (OWL) method~\cite{yin2023outlier} uses a nonuniform layerwise sparsity based on the distribution of outlier activations.
However, OWL relies on heuristics related to the presence of outliers~\cite{kovaleva2021bert,puccetti2022outliers,dettmers2022llm}.
This can lead to suboptimal performance in the absence of outliers, and this can make it difficult to achieve very aggressive levels of sparsity. 
For example,~\citet{yin2023outlier} shows that pruning LLMs to 80\% sparsity often significantly degrades the prediction performance of LLMs.

\begin{figure}[!th]
    \centering
    \includegraphics[width=\linewidth,keepaspectratio]{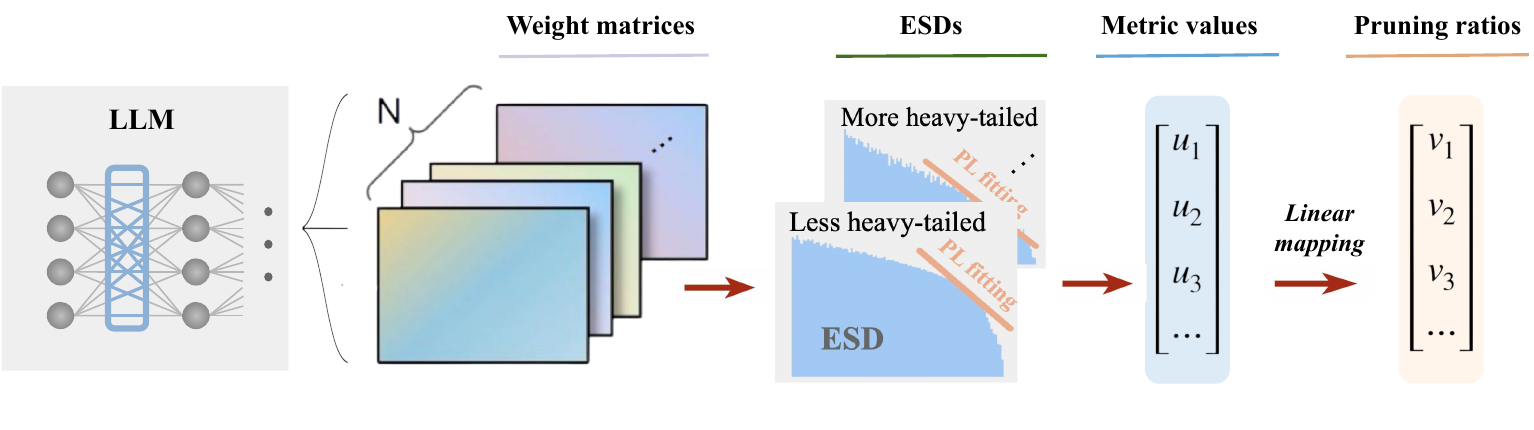} 
    \caption{\textbf{The pipeline diagram of \ourmethod}. Our post-training layer-wise pruning method involves the following steps: (\romannumeral1) Performing ESD analysis on all weight matrices of a base LLM and (\romannumeral2) employing PL fitting to derive the layer-wise metric values (that measures the HT exponent). Then, (\romannumeral3) using the layer-wise metric values, we assign layer-wise pruning ratios to each layer through a linear assignment function.}  \label{fig:teaser} 
\end{figure}

In developing a principled approach to allocate sparsity budgets across layers, we draw inspiration from Heavy-Tailed Self-Regularization (HT-SR) Theory~\cite{martin2020predicting_NatComm,MM21a_simpsons_TR,martin2018implicit_JMLRversion,martin2019traditional,martin2017rethinking,martin2020heavy,yang2023test,zhou2023temperature}. 
HT-SR theory analyzes the weight matrices of models to derive quantities (related to the shape of the weight matrix eigenspectrum), which help characterize model capacity and quality.
Applications of HT-SR to model selection~\cite{martin2019traditional, martin2020heavy, MM21a_simpsons_TR,martin2020predicting_NatComm,yang2023test} 
and layer-wise adaptive training~\cite{zhou2023temperature} demonstrate the effectiveness of the theory in estimating model and layer quality.
Furthermore, in the context of pruning memory/computation-efficient LLMs, using this kind of low-cost weight analysis is advantageous because it requires no data or gradient backpropagation. 

Our study consists of two main parts.
In the first part, we evaluate the effectiveness of various weight matrix-based metrics for allocating layer-wise sparsity. 
Our primary finding indicates that \emph{shape metrics} generally outperform \emph{scale metrics} in determining layer importance for pruning. 
Shape metrics capture the shape properties of the empirical spectral densities (ESDs) of layer weight matrices, whereas scale metrics, like matrix norms, reflect the size or scale of ESDs. 
This offers a novel perspective since shape metrics are less frequently used in the literature than scale metrics, which have been used to create regularizers~\cite{yoshida2017spectral} and inform pruning~\cite{han2015learning}.

In the second part of the paper, we introduce a theoretically principled layer-wise sparsity allocation method, \ourmethod, based on metrics that quantify a unique heavy-tailed (HTed) shape of ESDs.
According to HT-SR theory, well-trained models display strong correlations among the weight matrix elements, leading to HT structures in layer weight matrices' ESDs~\cite{martin2019traditional, martin2020heavy, MM21a_simpsons_TR}. 
Moreover, layers with more pronounced HT properties are typically better trained than others. 
We quantify the HT properties by fitting a power law (PL) distribution~\cite{alstott2014powerlaw, clauset2009power} to ESD and using the PL exponent as the HT metric \ALPHAHILL.\footnote{Following \citet{zhou2023temperature}, we use a biased estimator, called \ALPHAHILL, to replace the commonly used MLE estimator of the PL exponent~\cite{alstott2014powerlaw, clauset2009power,martin2018implicit_JMLRversion} for allocating layer-wise sparsity ratios. This estimator proves to have a smaller variance and thus is easier to integrate with training models that require repeated estimation of the HT exponent.}
The principle of \ourmethod is to allocate less sparsity to more well-trained (more HTed) layers, as indicated by lower \ALPHAHILL values, thereby preserving their quality during pruning.
Figure~\ref{fig:teaser} illustrates the pipeline of \ourmethod.

We conducted a comprehensive empirical evaluation to assess the generalizability of \ourmethod in LLM pruning. This evaluation involved comparisons with various baseline methods, integration with existing techniques, and testing across different architectures. We also conducted sanity checks to ensure that \ourmethod indeed reduces the variation of \ALPHAHILL values across layers.
Our key contributions are summarized below.
\begin{itemize}
[noitemsep,topsep=0pt,leftmargin=*,after=,before=]
    \item 
    This paper is the first to study principled layer-wise sparsity allocation from the HT-SR perspective. 
    We systematically evaluate multiple weight matrix-based metrics to compute sparsity based on their effectiveness in estimating layer quality, discovering an interesting finding: shape metrics outperform scale metrics in allocating sparsity, despite the latter being more commonly used.

    \item 
    We introduce a novel sparsity allocation method, \ourmethod, inspired by HT-SR theory, which demonstrates superior performance in LLM pruning. 
    This method assigns sparsities based on the heavy-tailed shape of ESDs in layer weight matrices, a previously unexplored concept. 
    Our empirical evaluations span a range of LLM architectures, including the LLaMA V1-3 families~\cite{touvron2023llama, touvron2023llama2}, OPT families~\cite{zhang2023opt}, Vicuna-7B~\cite{zheng2024judging}, and Mistral-7B~\cite{jiang2023mistral}. 
    The results show that \ourmethod outperforms OWL~\cite{yin2023outlier}, the current SOTA non-uniform sparsity allocation method for LLMs, reducing perplexity by 304.31 and achieving an average accuracy gain of 4.6\% over 7 zero-shot tasks at 80\% sparsity, while providing a 3.06$\times$ end-to-end speedup on CPUs for LLaMA-7B on the DeepSparse~\cite{deepsparse} inference engine. 
    \ourmethod also outperforms six layer-wise allocation methods, including global thresholding~\cite{frankle2018lottery}, ER~\cite{mocanu2018scalable}, ER-Plus~\cite{liu2022unreasonable}, LAMP~\cite{lee2020layer}, rank selection~\cite{kuzmin2019taxonomy, el2022data}, and layer-wise error thresholding~\cite{ye2020good, zhuang2018discrimination}.
    
    \item
    \ourmethod provides layer-wise budget allocation (e.g., sparsity), demonstrating remarkable generalizability and can be integrated with multiple LLM compression techniques to enhance performance.
    This includes unstructured pruning (Wanda~\cite{sun2023simple}, SparseGPT~\cite{frantar2023sparsegpt}), with or without fine-tuning~\cite{hu2021lora}, semi-structured (DominoSearch~\cite{sun2021dominosearch}), structured pruning (LLMPruner~\cite{ma2023llm}, OSSCAR~\cite{meng2024osscar}), mixed-precision quantization~\cite{tang2022mixed}. Additionally, we have extended this method to large Computer Vision (CV) architectures, such as Vision Transformers (ViT)~\cite{dosovitskiy2020image}, and ConvNext~\cite{liu2022convnet}. In this case, OWL significantly underperforms \ourmethod due to the lack of outlier features. This indicates HT metrics are more general than outlier metrics.

    \item 
    \ourmethod is theoretically driven, and its improvements can be interpreted by HT-SR metrics. 
    We demonstrate that model performance correlates with model-wise and layer-wise changes in \ALPHAHILL before and after pruning. 
    Furthermore, compared to baseline pruning methods, \ourmethod not only improves model performance but also achieves a lower mean of layer-wise \ALPHAHILL. 
    HT-SR theory suggests that \ourmethod preserves the quality of model layers ``on average'', minimizing the damage caused by pruning.

\end{itemize}

\ifisTR
\section{Related work}\label{sec:related_work}

\paragraph{Pruning.} 
Removing weights or connections in a trained neural network (NN) to generate an efficient, compressed model has a long history~\cite{lecun1990optimal, mozer1988skeletonization, janowsky1989pruning,mocanu2018scalable}.
Modern NNs are frequently over-parameterized~\cite{wang2020rethinking, bhojanapalli2021leveraging}, and thus removing redundancies improves computation and memory efficiency.
A common approach is weight-magnitude-based pruning \cite{han2015learning}, which zeros out connections with weights smaller than a specified threshold. 
However, when it comes to pruning LLMs~\cite{brown2020language, touvron2023llama}, progress has been limited. 
Conventional pruning typically requires a round of retraining to restore performance~\cite{blalock2020state}, which can be challenging for LLMs. 
To address the difficulty in retraining, researchers have developed specially tailored pruning algorithms for LLMs. 
For example, ~\citet{ma2023llm} explored sparse structured LLMs, using Taylor pruning to remove entire weight rows, followed by LoRA fine-tuning. 
More recent research has shifted towards unstructured pruning without the need for fine-tuning, showing substantial advancements. 
In particular, SparseGPT~\cite{frantar2023sparsegpt} uses the Hessian inverse for pruning and subsequent weight updates to reduce the reconstruction error of dense and sparse weights, while Wanda~\cite{sun2023simple} uses a criterion that incorporates weight magnitudes and input activations, in order to preserve outlier features~\cite{kovaleva2021bert,puccetti2022outliers,dettmers2022llm}. 
Our work allocates parameters in a more theoretically-principled manner, enabling pruning LLMs to higher sparsity levels.

\paragraph{Layerwise sparsity budgets.} 
Although layerwise sparsity has been widely studied in pre-LLM pruning~\citep{mocanu2018scalable,evci2020rigging,liu2022unreasonable,gale2019state,lee2020layer}, relatively little attention has been devoted to determining the pruning ratios for each layer in LLMs. 
(Interestingly, this layer-wise approach has been applied to model quantization~\cite{kim2023squeezellm,shen2020q,dong2019hawq}.)
\citet{frantar2023sparsegpt} and \citet{sun2023simple} apply a uniform pruning ratio across all layers, and
\citet{yin2023outlier} computes the sparsity budgets using the outlier ratio observed within each layer's token feature distribution.
Existing work on sparsity budgets has generally used heuristics, such as different forms of size or scale metrics (such as norm-based metrics), to determine sparsity budgets per layer.
For instance, ABCPruner~\cite{lin2020channel} reduces the number of combinations of layer sparsities to search over, but it still requires training to determine the empirical validity of its suggested layer sparsities.
\citet{lee2020layer} modifies Magnitude-based pruning by rescaling the importance scores in a layer by a factor dependent on the magnitude of surviving connections in that layer. 
However, these methods are suboptimal in allocating layerwise sparsities for pruning LLMs. 
Recently, Outlier Weighed Layerwise sparsity (OWL)~\cite{yin2023outlier} designs a nonuniform layerwise sparsity based on the distribution of outlier activations in LLMs. 
However, OWL heuristically relies on the emergence of outliers, and this can lead to suboptimal performance when outliers are absent from models.
Our work uses HT-SR shape metrics such as \ALPHAHILL to predict layer importance, and it allocates parameters in a more theoretically principled manner, allowing pruning LLMs to higher sparsity levels than has ever been achieved before.

\else
We provide an overview of related work in Appendix~\ref{sec:related_work}.
\fi

\section{Background and Setup}

\subsection{Notation}
Consider a NN with $L$ layers, $\mathbf{W}_i$ is one of the weight matrices extracted from the $i$-th layer with shape $m \times n$ ($m\ge n$). 
We note that the ``layer'' used in this work refers to the transformer block (layer), and each block contains multiple weight matrices, such as the attention layer weight matrix, and projection layer weight matrix.
The correlation matrix $\mathbf{X}_i=\mathbf{W}_i^\top \mathbf{W}_i$ is an $n \times n$ symmetric matrix, and the ESD of $\mathbf{X}_i$ is formulated as
\begin{align}
    \mu_{\mathbf{X}_i}:=\frac{1}{n} \sum_{j=1}^n \delta_{\lambda_j\left(\mathbf{X}_i\right)}
\end{align}
where $\lambda_1\left(\mathbf{X}_i\right) \leq \ldots \leq \lambda_n\left(\mathbf{X}_i\right)$ are the eigenvalues of $\mathbf{X}_i$ and $\delta$ is the Dirac delta function. The ESD is a probability measure, which can be viewed as a distribution of the eigenvalues of $\mathbf{X}_i$.

\subsection{HT-SR theory and metrics}\label{sec:def-metric}

Here, we provide a brief overview of HT-SR theory. HR-SR theory originated as a semi-empirical theory, with early seminal work~\cite{martin2019traditional, martin2020predicting_NatComm} examining the empirical spectral density (ESD) of weight matrices, specifically the eigenspectrum of the correlation matrix $\mathbf{X}_i=\mathbf{W}_i^\top \mathbf{W}_i$. This research found that the structures of the ESDs strongly correlate with training quality. 
These findings are rooted in statistical physics and Random Matrix Theory~\cite{couillet_liao_2022}, as detailed in Table 1 of \citet{martin2019traditional}. 
It is well-known~\cite{wang2024spectral, couillet_liao_2022} that spikes in ESD represent ``signals,'' while the bulk represents noise, which follows the Marchenko-Pastur law. 
In the theoretical setting of \citet{wang2024spectral}, the signal or the spike aligns with ground-truth features from the teacher model, and that corresponds to increased correlations in weight elements. 
Furthermore, \citet{kothapalli2024crafting} show that heavy tails in ESD originate from the interaction between spikes and bulk, which can be quantified precisely using recent advances in the free-probability theory~\cite{landau2023singular}, and the interaction characterizes the ``bulk-decay'' phase in the five-plus-one phase model in \citet{martin2019traditional}, a critical phase between classical ``bulk+spike'' model and heavy-tail models.

To quantify the structure of ESDs, HT-SR theory provides several metrics, collectively known as HT-SR metrics. These metrics are typically categorized into two groups: scale metrics and shape metrics.

\textbf{Scale metrics.}
Scale metrics refer to those obtained from measuring various norms of weight matrices. 
As demonstrated empirically in \citet{yang2023test}, these metrics are often strongly correlated with the generalization gap (which is the gap between training and test performance), instead of the quality of the models. 
In this paper, we mainly study two scale metrics, \NORM and \SPECTRALNORM. 
The \NORM metric is calculated by the squared Frobenius norm of the weight matrix $ \left \| \mathbf{W} \right \| ^2_F $; and the \SPECTRALNORM metric can be calculated by the square of the spectral radius of the weight matrix $\left \| \mathbf{W} \right \|_2^2$.

\textbf{Shape metrics.}
Drawing analytic methods from Random Matrix Theory, HT-SR work analyzes poorly trained and well-trained models and concludes that the performance of these models usually correlates with shapes emerging in their ESDs, such as ``bulk+spike'' shape or ``heavy-tailed'' shape.
The metrics used to characterize these ESD shapes are called shape metrics, and we mainly studied four of them: \ALPHAHILL, \ALPHAHAT, \STABLERANK, and \ENTROPY.
\ALPHAHILL is the main metric used in our method, and we define it in Section~\ref{sec:ht-metric}.
The definitions of other shape metrics (including \ALPHAHAT, \STABLERANK, \ENTROPY) can be found in Appendix~\ref{sec:supp-metrics}.

\section{Alpha-Pruning}

In this section, we first outline the motivation behind \ourmethod, followed by an introduction to the layer-wise importance metric, \ALPHAHILL, and the sparsity allocation function, as shown in Figure~\ref{fig:teaser}. 
Our empirical analysis reveals that shape metrics generally outperform scale metrics in guiding sparsity allocation with the same function. 
Notably, \ALPHAHILL, the shape metric used in \ourmethod, achieves the best results in preliminary evaluations.

\subsection{Rationale}
HT-SR theory, introduced in Section~\ref{sec:def-metric}, examines the ESD of weight matrices and finds a strong correlation between heavy-tailed structures in the ESD and training quality.
It suggests that heavy-tailed structures emerge from feature learning, where useful correlations are extracted during optimization. Layers with more heavy-tailed ESDs tend to capture more signals, indicating better training quality. 
Inspired by these findings, we propose to assign sparsity based on the heavy-tailed properties of each layer’s ESD. 
Layers with more heavy-tailed ESDs, which contain more learned signals, are assigned lower sparsity, while layers with light-tailed ESDs are assigned higher sparsity. 
In practice, the heavy-tailed structure is measured by fitting a PL distribution to the ESD, and extracting the PL exponent $\alpha$ as the indicator. This is why our method is named \ourmethod.

\subsection{Estimating layer quality by HT metric}\label{sec:ht-metric}

\ourmethod relies on estimating the layer quality based on the HT characteristic of the layer ESDs, which is quantified by HT metric \ALPHAHILL.
Given an ESD $\mu_{\mathbf{X}_i}$ of a weight matrix's correlation matrix, we fit a PL density function $p(\lambda)$ on it, taking values within an interval ($\lambda_\text{min}$, $\lambda_\text{max}$), formally defined as:
\begin{equation}\label{eqn:ALPHA}
    p(\lambda) \propto \lambda^{-\alpha}, \lambda_\text{min} < \lambda < \lambda_\text{max}.
\end{equation}

The estimated exponent $\alpha$ is then used as a metric to characterize the HT extent of the ESD, with a lower value means more HTed. 
We estimate the PL coefficient using the Hill estimator~\cite{hill1975simple, xiao2023heavy,zhou2023temperature}, and we refer to it as the \ALPHAHILL metric.
The Hill estimator is defined as: 
\begin{equation}
\label{eqn:hill_estimator}
\ALPHAHILL= 1+\frac{k}{(\sum_{i=1}^k \ln\frac{\lambda_{n-i+1}}{\lambda_{n-k}})},
\end{equation}
where $\lbrace \lambda_{i} \rbrace_{i=1}^n$ is sorted in ascending order, and $k$ is a tunable parameter that adjusts the lower eigenvalue threshold $\lambda_\text{min}$ for (truncated) PL estimation. 
We adopt the Fix-finger method~\cite{yang2023test} to select the $k$, which sets $k$ such that $\lambda_\text{min}$ aligns with the peak of the ESD.
Note that \ALPHAHILL and other scale/shape metrics are calculated for each weight matrix individually.

\subsection{Allocating sparsity based on the layer quality}
\label{sec:allocation}

\ourmethod allocates sparsity for each layer (transformer block) by using a mapping function $\phi: \mathbb{R}^L \rightarrow \mathbb{R}^L$ to map a sequence of layer quality measures $\mathbf{q} = (q_1, q_2,..,q_L)$ into corresponding sparsities $\phi(\mathbf{q})$. 

\begin{equation}\label{eqn:sparsity-all}
\phi(\mathbf{q})_i = \eta \left[ \frac{q_i - q_{\text{min}}}{q_{\text{max}} - q_{\text{min}}} (s_2 - s_1)  + s_1 \right].
\end{equation}

Here, $\phi(\mathbf{q})_i$ represents the $i$-th element of the resulting vector $\phi(\mathbf{q})$, $q_i$ represents the $i$-th element of the input vector $\mathbf{q}$, and $q_\text{min}$, $q_\text{max}$ represent the minimum and maximum values of $\mathbf{q}$.
The normalization factor $\eta$ adjusts the sparsity levels to achieve the target global sparsity $S$. 
Each layer's sparsity is normalized within the interval $\left[\eta s_1, \eta s_2 \right]$. $\eta$ is calculated using the equation $\sum_{i=1}^L \phi(\mathbf{q})_i  d_i = S \cdot \sum_{i=1}^L d_i$, in which $d_i$ is the number of parameters of $\mathbf{W}_i$. 
Both sides of the equation represent the total number of remaining parameters.
The $(s_1, s_2)$ are tunable hyperparameters that adjust the non-uniformity of the sparsity distribution. We note that sparsity allocation is executed on a per-block basis, averaging the HT-SR metric across all matrices within a block to determine $q_i$.
This design is supported by an ablation study, presented in Appendix~\ref{app:ablation-allocation}, which shows that it yields superior performance over a per-matrix allocation.
Hyperparameter settings for all experiments are provided in Appendix~\ref{sec:supp-hyper}.

\subsection{Shape vs. scale metrics for sparsity allocation}
In this study, we evaluated various HT-SR metrics, as defined in Section~\ref{sec:def-metric}, to assess their effectiveness in estimating layer quality $\mathbf{q}$ for computing layer-wise sparsity. 
Our preliminary experiments involved pruning the LLaMA-7B model to 70\% sparsity and assessing its performance through WikiText perplexity and accuracy across seven zero-shot tasks, with results presented in Table~\ref{tb:shape-scale-llm}.
We applied each HT-SR metric in conjunction with three intra-layer pruning techniques (which only determine which matrix elements to prune), such as Magnitude, Wanda, and SparseGPT, to thoroughly evaluate their efficacy.
Further experiments on Vision Transformers (ViT) are described in Appendix~\ref{sec:supp-vit}. 
Across all tests, shape metrics consistently outperformed scale metrics in assigning layer-wise sparsities. 
This finding suggests that shape metrics are more robust and yield more reliable predictions of layer quality, which extends previous research~\cite{yang2023test, zhou2023temperature, martin2020predicting_NatComm} on estimating model quality.
Notably, the shape metric \ALPHAHILL that focuses on estimating the HT shape, proved to be the most effective.
Consequently, we have adopted \ALPHAHILL as the primary metric in our proposed method, \ourmethod.

\begin{table*}[!thb]
    \centering
    \resizebox{1.0\textwidth}{!}{
    \begin{tabular}{c|ccc|ccc}
        \toprule
        Metric used for & \multicolumn{3}{c|}{Perplexity on WikiText ($\downarrow$)} & \multicolumn{3}{c}{Average accuracy on 7 zero-shot tasks ($\uparrow$)} \\
        layerwise pruning ratios & Magnitude & Wanda & SparseGPT & Magnitude & Wanda & SparseGPT \\
        \midrule 
        Uniform & 48419.13 & 85.77 & 26.30 & 32.30 & 36.73 & 41.52  \\
        \midrule
        \cellcolor{smallColorT} \NORM & 30136.37 & 59.82 & 24.95 & 33.23 & 37.62 & 43.67 \\
        \cellcolor{smallColorT} \SPECTRALNORM & 48073.99 & 246.84 & 29.01 & 32.81 & 33.15 & 41.14 \\
        \midrule
        \cellcolor{largeColorT} \ENTROPY & 3716.07 & 41.15 & 22.02 & 33.58 & 39.39 & 43.53 \\
        \cellcolor{largeColorT} \STABLERANK & 851.65 & 41.24 & 24.91 & 34.91 & 39.74 & 42.97 \\
        \cellcolor{largeColorT} \ALPHAHAT & 1256.02 & 27.60 & 20.02 & 33.48 & 43.57 & 43.83 \\
        \rowcolor{LightBlue}
        \cellcolor{largeColorT} \ALPHAHILL & \bf{231.01} & \bf{23.86} & \bf{18.54}  & \bf{35.67} & \bf{44.42} & \bf{45.48} \\
        \bottomrule
    \end{tabular}
    }
    \caption{\textbf{Evaluating shape metrics versus scale metrics on allocating layerwise sparsities on LLMs.} \colorbox{largeColorT}{\textit{Shape metrics}} are obtained from the shapes
    of the ESDs. \colorbox{smallColorT}{\textit{Scale metrics}} are norm-based metrics measuring the scale of weights matrices (which can also be obtained from the ESD). The results are conducted on LLaMA-7B at 70\% sparsity. We show WikiText validation perplexity and average accuracy on seven different zero-shot tasks as evaluation metrics. We observe that shape metrics outperform scale metrics and \ALPHAHILL performs the best.} \label{tb:shape-scale-llm} 
\end{table*}

\section{Empirical results}
In this section, we evaluate the performance, generalizability, and interpretability of \ourmethod.
Section~\ref{sxn:empirical_setup} outlines our experimental setup. In Section~\ref{sec:main-compare-baseline}, we evaluate \ourmethod's performance by comparing it to the SOTA method OWL and five other baseline methods, and we analyze the efficiency of LLMs pruned by \ourmethod using practical metrics such as FLOPs and latency.
Section~\ref{sec:main-corro} evaluates the generalizability of \ourmethod by integrating it with various LLM compression techniques including post-pruning fine-tuning, semi-structured pruning, structured pruning, and mixed-precision quantization. Additionally, we extend its application to CV tasks.
Section~\ref{sec:main-analysis} provides an analysis of the layer-wise sparsities and the \ALPHAHILL distribution to further elucidate the effectiveness and implications of \ourmethod.

\subsection{Experimental setup}
\label{sxn:empirical_setup}

{\bf Models and Evaluation.} 
We evaluate \ourmethod on the three most widely adopted LLM model families: LLaMA 7B/13B/30B/65B~\cite{touvron2023llama}, LLaMA-2 7B/13B/70B~\cite{touvron2023llama2}, OPT 125M/350M/2.7B/6.7B, and other advanced LLMs: LLaMA-3-8B, Vicuna-7B, Mistral-7B.
Our evaluation protocol aligns with established methodologies for LLM pruning~\cite{xiao2023smoothquant}, including assessments of language modeling proficiency and zero-shot capabilities.
Specifically, we evaluate the perplexity on the held-out WikiText~\cite{merity2016pointer} validation set, and use seven tasks, including BoolQ~\cite{clark2019boolq}, RTE~\cite{wang2018glue}, HellaSwag~\cite{zellers2019hellaswag}, WinoGrande~\cite{sakaguchi2021winogrande}, ARC Easy and Challenge~\cite{clark2018think} and OpenbookQA~\cite{mihaylov2018can} for downstream zero-shot evaluation~\cite{eval-harness}.

\paragraph{Baselines.} 
We apply the layer-wise sparsities determined by \ourmethod to three LLM pruning methods, including Magnitude~\cite{han2015learning}, SparseGPT~\cite{frantar2023sparsegpt} and Wanda~\cite{sun2023simple}.   
Magnitude-based pruning is a simple and strong baseline in which weights are discarded based on their magnitudes. Wanda and SparseGPT are two strong LLM pruning baselines due to their capability to sustain reasonable performance even at relatively high sparsity levels (around 50\%). 
All these methods originally use uniform layerwise sparsity. 
We incorporate \ourmethod directly into these baselines, and we demonstrate that this results in improved performance. 
Besides, we also compare \ourmethod with OWL~\cite{yin2023outlier}, a recently proposed non-uniform LLM pruning method and six layer-wise pruning methods, including global thresholding~\cite{frankle2018lottery}, ER~\cite{mocanu2018scalable}, ER-Plus~\cite{liu2022unreasonable}, LAMP~\cite{lee2020layer}, rank selection~\cite{kuzmin2019taxonomy, el2022data}, and layer-wise error thresholding~\cite{ye2020good, zhuang2018discrimination}.

\subsection{Main results}
\label{sec:main-compare-baseline}
{\bf Language Modeling.}
In Table~\ref{tb:lm-main-results}, we report the perplexity of the pruned LLaMA and LLaMA-2 models at 70\% sparsity. We provide results for more sparsity levels in Figure~\ref{fig:perplexity-comp} and Appendix~\ref{sec:supp-baselines}.
\ourmethod, as a general layerwise sparsity method, consistently demonstrates performance improvements when used in conjunction with various pruning methods. 
For example, in the case of LLaMA-7B with a sparsity of 70\%, \ourmethod produces sparse networks with a perplexity of 231.01, significantly outperforming the Magnitude-based pruning baseline of 48419.13.
Notably, when applied to Wanda and SparseGPT, two robust LLM pruning methods, \ourmethod still achieves substantial perplexity reductions, evidenced by a decrease of 61.91 for Wanda and 7.76 for SparseGPT, in the case of LLaMA-7B with a sparsity of~70\%.

\begin{table*}[!thb]
    \centering
    \resizebox{0.9\textwidth}{!}{
    \begin{tabular}{c|c|cccc|ccc}
        \toprule
         & & \multicolumn{4}{c|}{LLaMA}  & \multicolumn{3}{c}{LLaMA-2} \\
        \multirow{-2}{*}{Method} & \multirow{-2}{*}{Layerwise sparsity} & 7B & 13B & 30B & 65B & 7B & 13B & 70B \\
        \midrule 
        Dense model & - & 5.68 & 5.09 & 4.77 & 3.56 & 5.12 & 4.57 & 3.12  \\ 
        \midrule
        & Uniform & 48419.13 & 84527.45 & 977.76 & 46.91 & 49911.45 & 214.04 & 1481.95   \\
        & OWL & 19527.58 & 11464.69 & 242.57 & \bf{15.16} & 59176.42 & 57.55 & 17.18 \\
        \rowcolor{LightBlue} \cellcolor{white} \multirow{-3}{*}{Magnitude} &  Ours & \bf{231.01} & \bf{2029.20} & \bf{62.39} & 16.01  & \bf{8900.32} & \bf{31.89} & \bf{15.27} \\
        \midrule 
        & Uniform & 85.77 & 54.03 & 17.35  & 15.17  & 74.26 & 45.36 & 10.56  \\
        & OWL & 24.57 & 17.17 & 10.75 & 8.61 & 30.38 & 20.70 & 8.52  \\
        \rowcolor{LightBlue} \cellcolor{white} \multirow{-3}{*}{Wanda} & Ours & \bf{23.86} & \bf{14.21} & \bf{9.68} & \bf{7.86} & \bf{28.87} & \bf{14.16} & \bf{7.83} \\
        \midrule 
        & Uniform & 26.30 & 18.85 & 12.95 & 10.14  & 27.65 & 19.77 & 9.28 \\
        & OWL & 19.49 & 14.55 & 10.28 & 8.28 & 20.40 & 15.27 & 7.65 \\
        \rowcolor{LightBlue}
        \cellcolor{white} \multirow{-3}{*}{SparseGPT} &  Ours  & \bf{18.54}  & \bf{12.81} & \bf{9.77} & \bf{7.83}  & \bf{19.34} & \bf{12.20} & \bf{7.57}  \\
        \bottomrule
    \end{tabular}
    }
    \caption{WikiText validation perplexity for pruned LLaMA and LLaMA-2 models at 70\% sparsity. Our method (\ourmethod) is compared to uniform layerwise sparsity and OWL, each paired with magnitude-based pruning, Wanda, and SparseGPT. Lower perplexity indicates improved model performance.}
    \label{tb:lm-main-results}
\end{table*}

\paragraph{Zero-shot tasks.}
We conducted empirical evaluations to determine the zero-shot ability of pruned LLMs on diverse zero-shot downstream tasks with prompting. 
The results are shown in Table~\ref{tb:zeroshot-main-results}, where we show the mean zero-shot accuracy on 7 zero-shot tasks of pruned LLaMA and LLaMA-2 models at sparsity of 70\%. 
\ourmethod consistently improves accuracy across all settings. 
For example, \ourmethod achieves an average accuracy gain of 8.79, 6.05, and 2.61 over 7 tasks and 7 models compared to Magnitude, Wanda and SparseGPT alone, respectively. 
These results highlight the promise of \ourmethod for more challenging zero-shot downstream tasks.

\begin{table*}[!thb]
    \centering
    \resizebox{0.8\textwidth}{!}{
    \begin{tabular}{c|c|cccc|ccc}
        \toprule
         & & \multicolumn{4}{c|}{LLaMA}  & \multicolumn{3}{c}{LLaMA-2} \\
        \multirow{-2}{*}{Method}  & \multirow{-2}{*}{Layerwise sparsity} & 7B & 13B & 30B & 65B & 7B & 13B & 70B \\
        \midrule 
        Dense model & - & 60.08 & 62.59 & 65.47 & 67.05 & 59.71 & 63.05 & 67.06  \\ 
        \midrule
        & Uniform & 32.30 & 34.95 & 32.39 & 45.04 & 33.38 & 33.56 & 43.65   \\
        & OWL & 33.57 & 36.86 & 33.88 & 53.42 & 34.17 & 36.98 & 50.49 \\
        \rowcolor{LightBlue} \cellcolor{white} \multirow{-3}{*}{Magnitude} &  Ours & \bf{35.67} & \bf{38.23} & \bf{42.56} & \bf{55.22} & \bf{37.86} & \bf{44.53} & \bf{50.72}  \\
        \midrule 
        & Uniform & 36.73 & 38.90 & 51.07 & 54.90 & 34.66 & 37.16 & 56.72  \\
        & OWL & 43.41 & 45.91 & 52.38 & 57.34 & 40.46 & 45.04 & 57.19 \\
        \rowcolor{LightBlue}  \multirow{-3}{*}{Wanda} &  Ours & \bf{44.42} & \bf{47.48} & \bf{54.48} & \bf{58.93} & \bf{41.98} & \bf{47.21} & \bf{57.63} \\
        \midrule 
        & Uniform & 41.52 & 45.67 & 52.75 & 57.81 & 41.84 & 45.17 & 59.68  \\
        & OWL & 44.65 & 47.61 & 53.16 & 58.25 & \bf{44.96} & 47.94 & 59.20 \\
        \rowcolor{LightBlue} \cellcolor{white} \multirow{-3}{*}{SparseGPT} &  Ours  & \bf{45.48} & \bf{48.41} & \bf{54.07} & \bf{59.72} & 44.68 & \bf{49.26} & \bf{60.23} \\
        \bottomrule
    \end{tabular}
    }
    \caption{Comparison of mean zero-shot accuracies (\%) for pruned LLaMA and LLaMA-2 models at 70\% sparsity. We evaluate our method (\ourmethod) against uniform layerwise sparsity and OWL, each integrated with magnitude-based pruning, Wanda, and SparseGPT. Higher accuracy values indicate better zero-shot ability.} \label{tb:zeroshot-main-results} 
\end{table*}

\paragraph{More baseline comparison.} For allocating layerwise sparsity ratios, we compare \ourmethod with other allocation methods.
The experiments involve pruning the LLaMA-7B model and LLaMA-13B to various sparsities.
We use Wanda as the basic pruning method, with results presented in Figure~\ref{fig:perplexity-comp} and Figure~\ref{fig:accuracy-comp}.
Results indicate that \ourmethod significantly outperforms all baseline methods in relatively high-sparsity regimes. For LLaMA-7B and LLaMA-13B pruned to 80\% sparsity, \ourmethod reduces perplexity by 304.31 and 200.54 compared to OWL, respectively.
Achieving high levels of sparsity is crucial for unstructured sparsity to yield significant speedups on GPUs by leveraging existing sparse kernels. Sparse kernels such as Flash-LLM~\cite{xia2023flash} and Sputnik~\cite{gale2020sparse} have shown that unstructured sparsity outperforms dense computation in terms of performance, but only when sparsity levels reach 60\% and 71\%, respectively.
The importance of achieving high sparsity is further substantiated in the following section, which demonstrates that high sparsity levels facilitate significant end-to-end inference speedup. 
Additional comparisons with rank selection and layer-wise error thresholding, as well as results using SparseGPT as the pruning method and low sparsity results, can be found in Appendices~\ref{sec:supp-baselines}.

\begin{minipage}{\textwidth}
\begin{minipage}[t]{\textwidth}
\makeatletter\def\@captype{figure}
\centering
\includegraphics[width=0.75\linewidth]{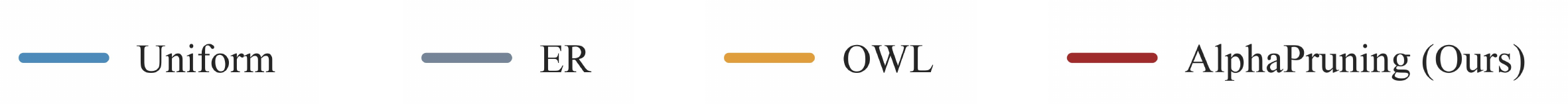}
\end{minipage}
\begin{minipage}[t]{0.63\textwidth}
\makeatletter\def\@captype{figure}
\centering
    \includegraphics[width=0.49\linewidth]{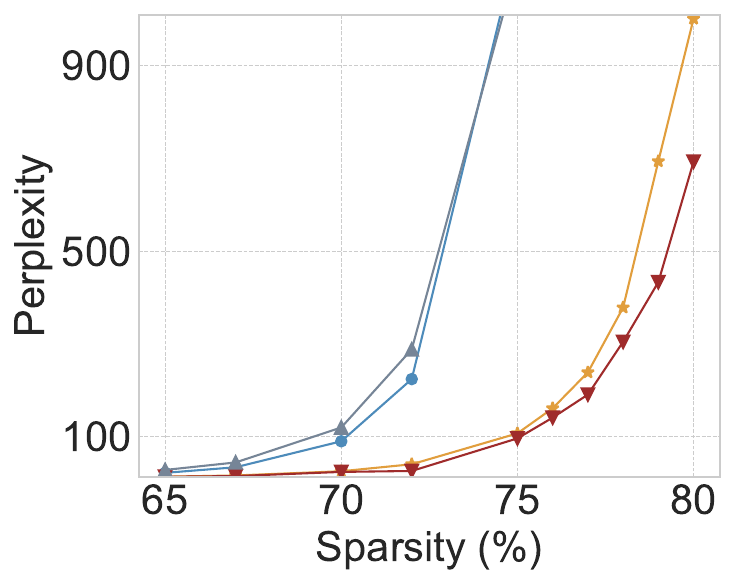}
    \includegraphics[width=0.49\linewidth]{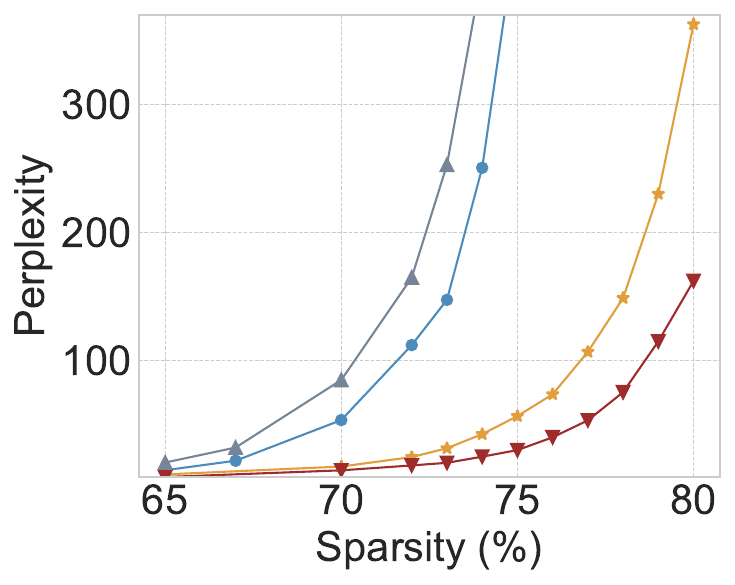}
    \caption{The WikiText validation perplexity for LLaMA-7B (left) and LLaMA-13B (right) pruned with different sparsities using Wanda.}\label{fig:perplexity-comp}
\end{minipage}
\hspace{0.02\textwidth}
\begin{minipage}[t]{0.3\textwidth}
\makeatletter\def\@captype{figure}
\centering
    \includegraphics[width=\linewidth]{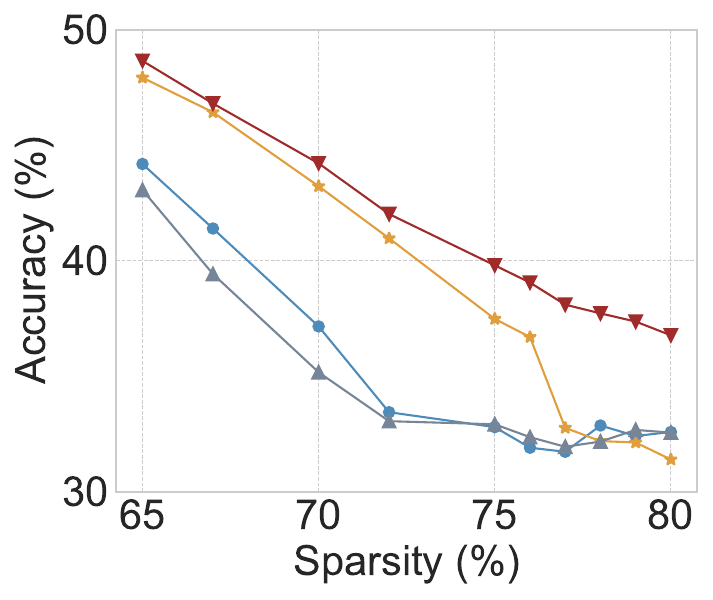} 
    \caption{Mean accuracy of 7 zero-shot tasks for LLaMA-7B pruned with different sparsities using Wanda.
    }\label{fig:accuracy-comp}
\end{minipage}
\end{minipage}

\paragraph{Efficiency measure.} 
To verify the sparse LLM pruned by our method can indeed achieve speedups when deployed on the CPU, we provide new results in Table~\ref{tb:speedup}.
We apply our method to Llama2-7B-Chat-hf, prune it to different sparsities, and then test its end-to-end decode latency using DeepSparse~\cite{kurtic2023sparse} inference engine on an Intel Xeon Gold 6126 CPU with 24 cores. 
The results indicate that when the global sparsity reaches 80\%, the speedup reaches 3.06$\times$.

\begin{table*}[!thb]
    \centering
    \resizebox{1.0\textwidth}{!}{
    \begin{tabular}{ccccccccccc}
        \toprule
        Sparsity & Dense & 10\% & 20\% & 30\% & 40\% & 50\% & 60\% & 70\% & 80\% & 90\% \\
        \midrule 
        Latency (ms) & 307.46 &	306.27 & 304.87 & 293.26 & 264.41 &	177.55 & 148.16 & 133.76 & 100.35 & 81.40 \\
        Throughput (tokens/sec) & 3.25 & 3.26 &	3.28 & 3.41 & 3.78 & 5.63 &	6.74 & 7.47 & 9.96 & 12.28 \\
        Speedup & 1.00x & 1.00x & 1.01x & 1.05x & 1.16x & 1.73x & 2.07x & 2.30x & 3.06x & 3.78x \\
        \bottomrule
    \end{tabular}
    }
    \caption{ End-to-end decode latency and speedup of \ourmethod measured on the DeepSparse inference engine.
    }\label{tb:speedup}
\end{table*}

In Appendix~\ref{sec:app-other-eff}, we evaluate the pruned LLM by efficiency metrics other than sparsity, such as FLOPs, Compared with uniform sparsity ratios, our approach is able to achieve better performance-FLOPs trade-off. 
In Appendix~\ref{sec:app-mini-spar}, we show that \ourmethod can control the minimum layer sparsity without losing the performance advantage to meet the hardware requirements of a memory-limited device.
In Appendix~\ref{sup:comp-complexity}, we report the runtime of \ourmethod to show that the computational overhead is reasonable. The computational complexity is not large because the most computation-intensive aspect of our method involves performing SVD decomposition on weight matrices.

\subsection{Corroborating results}
\label{sec:main-corro}
To demonstrate the generalizability of \ourmethod, we first evaluate if the performance of the model pruned by \ourmethod can be well recovered by fine-tuning. 
Then we apply \ourmethod to other LLM compression techniques (semi-structured, structured pruning, and quantization), and CV models.\looseness-1

\paragraph{Fine-tuning.}
We show the performance of LLMs pruned by \ourmethod can be well recovered by fine-tuning.
We investigate the parameter-efficient strategies for fine-tuning LLMs: LoRA fine-tuning~\cite{hu2021lora}. Fine-tuning is conducted on the C4 training set~\cite{raffel2020exploring} with the pre-training auto-regressive loss. 
The pruned mask is fixed during fine-tuning. The low-rank ($r$ = 8) adapter is applied to the query and value projection matrices in the attention layers. 
We fine-tune LLaMA-7B pruned by SparseGPT at various sparsities. 
Table~\ref{tb:corro-lora} summarizes the results for perplexity and mean zero-shot accuracy after fine-tuning pruned LLaMA-7B models. We can see the performance of pruned LLMs can be notably improved with very light LoRA fine-tuning. 
In Appendix~\ref{sec:app-fine-tuning}, we further compare \ourmethod with baselines. The results show that the advantages of \ourmethod don’t diminish after fine-tuning.

\vspace{2mm}
\begin{minipage}{\textwidth}
\begin{minipage}[t]{0.45\textwidth}
\makeatletter\def\@captype{table}
    \resizebox{1.0\textwidth}{!}{
    \begin{tabular}{cccccc}
        \toprule
        Method & Model & 60\% & 70\% & 80\% \\
        \midrule 
        Uniform & LLaMA-V3-7B & 23.35 & 126.57 & 880.57  \\
        \rowcolor{LightBlue}
        Ours & LLaMA-V3-7B & \bf{18.91} & \bf{101.73} & \bf{560.22} \\
        \midrule 
        Uniform & Vicuna-7B & 12.64 & 110.05 & 6667.68 \\
        \rowcolor{LightBlue}
        Ours & Vicuna-7B & \bf{11.14} & \bf{35.87} & \bf{1081.76} \\
        \midrule 
        Uniform & Mistral-7B & 11.47 & 60.08 & 353.62 \\
        \rowcolor{LightBlue}
        Ours & Mistral-7B & \bf{10.49} & \bf{39.39} & \bf{201.62} \\
        \bottomrule
    \end{tabular}
    }
    \caption{WikiText validation perplexity ($\downarrow$) of more LLMs pruned by uniform sparsity and our method combined with Wanda.
    }~\label{tb:advanced-llm}
\end{minipage}
\hspace{0.01\textwidth}
\begin{minipage}[t]{0.52\textwidth}
\makeatletter\def\@captype{table}
\resizebox{1.0\textwidth}{!}{
    \begin{tabular}{cccccc}
        \toprule
        Method & Sparsity & Samples & Perplexity ($\downarrow$) & Accuracy ($\uparrow$) \\
        \midrule
        -- & Dense & -- & 5.68 & 60.08 \\
        \midrule
        Uniform & 50\% & 20K & 6.86 & 54.85 \\
        \rowcolor{LightBlue}
        Ours & 50\% & 20K & \bf{6.79} & \bf{55.58} \\
        \midrule
        Uniform & 60\% & 20K & 8.32 & 51.81 \\
        \rowcolor{LightBlue}
        Ours & 60\% & 20K & \bf{8.23} & \bf{53.39} \\
        \midrule
        Uniform & 70\% & 30K & 11.21 & 49.00 \\
        \rowcolor{LightBlue}
        Ours & 70\% & 30K & \bf{10.95} & \bf{49.51} \\
        \midrule
        Uniform & 80\% & 100K & 20.11 & 39.32 \\
        \rowcolor{LightBlue}
        Ours & 80\% & 100K & \bf{18.53} & \bf{40.02} \\
        \bottomrule
    \end{tabular}
    }
    \caption{WikiText validation perplexity and zero-shot tasks accuracy of SparseGPT pruned LLaMA-7B at various sparsities after LoRA fine-tuning on C4 dataset samples.
    }\label{tb:corro-lora}
\end{minipage}
\end{minipage}

\begin{wrapfigure}[18]{r}{0.4\textwidth}
\centering
\includegraphics[width=1.0\linewidth,keepaspectratio]{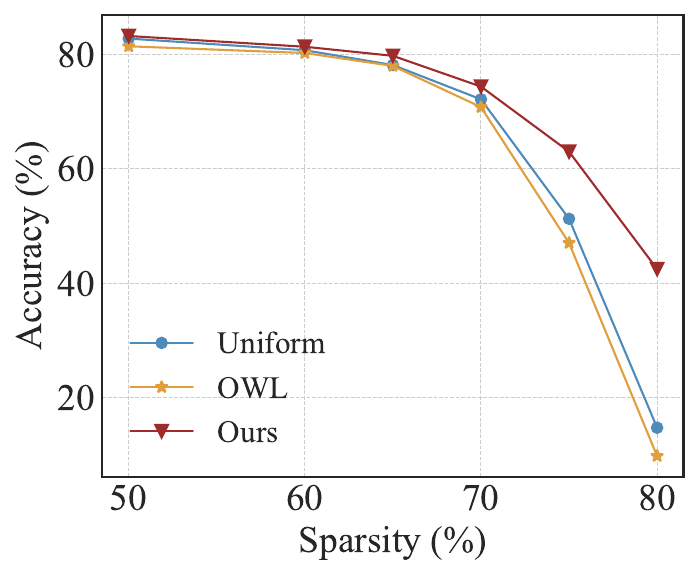} 
\caption{ImageNet-1K accuracy ($\uparrow$) of the sparse ConvNext model pruned to various sparsity levels by AlphaPruning and other baseline methods, without fine-tuning.
}
\label{fig:convnext}
\end{wrapfigure}

\paragraph{More LLM architectures.}
To demonstrate that the effectiveness of \ourmethod is robust across various more advanced LLMs, we also apply \ourmethod to LLaMA-3-7B, Vicuna-7B~\cite{zheng2024judging}, and Mistral-7B~\cite{jiang2023mistral}. 
The results in Table~\ref{tb:advanced-llm} show that, as a general method, \ourmethod can consistently achieve performance improvement across different architectures. Results for OPT families can be found in Appendix~\ref{other-llm-family}.

\paragraph{Integration with other compression techniques.} 
To demonstrate the generalizability of our non-uniform layerwise sparsity, we integrated \ourmethod with three prominent LLM compression methods: N:M sparsity, structured pruning, and quantization. 
In Appendix~\ref{sec:supp-structured}, we examine a mixed N:8 sparsity setup using DominoSearch~\cite{sun2021dominosearch} as well as combine \ourmethod with structured pruning methods LLM-Pruner~\cite{ma2023llm} and OSSCAR~\cite{meng2024osscar}. 
In Appendix~\ref{sec:app-quantization}, we merge our method with mixed-precision quantization as ~\cite{tang2022mixed}. 
Across these configurations, \ourmethod consistently enhances the performance of baseline methods.

\paragraph{Vision models.} 
To illustrate \ourmethod in a broader context, we study how it performs against other methods for determining layerwise sparsity on CV tasks, where non-uniform layerwise sparsity has been widely used. We consider two modern CV architectures: ViT~\cite{dosovitskiy2020image} and ConvNext~\cite{liu2022convnet}. 
We adopt Wanda as the pruning approach and compare \ourmethod with uniform layerwise sparsity and OWL.
We present the results on ConvNext in Figure~\ref{fig:convnext} and provide more results on DeiT and ViT models in Appendix~\ref{sec:app-cv-results}. These results affirm that \ourmethod effectively allocates layerwise sparsities in CV tasks as well, where OWL significantly underperforms \ourmethod due to the lack of outlier features. This indicates HT metrics are more general than outlier metrics.

\subsection{Analyzing LLM pruning via HT-SR perspective}\label{sec:main-analysis}

First, we investigate the distribution of layerwise sparsities allocated by the heavy-tailed metric \ALPHAHILL.
Then, we study how the \ALPHAHILL metric as a measure of model quality (based on HT-SR theory) changes before and after pruning.
In particular, we show that the proposed method, \ourmethod, effectively controls the damage of pruning to model quality, resulting in a more favorable distribution (lower mean) of \ALPHAHILL among the model layers compared to the baseline pruning method. Our primary focus is on LLaMA, with additional analyses on other LLMs provided in Appendix~\ref{more-analysis}.

\begin{wrapfigure}{r}{0.4\textwidth}
\centering
\includegraphics[width=1.0\linewidth,keepaspectratio]{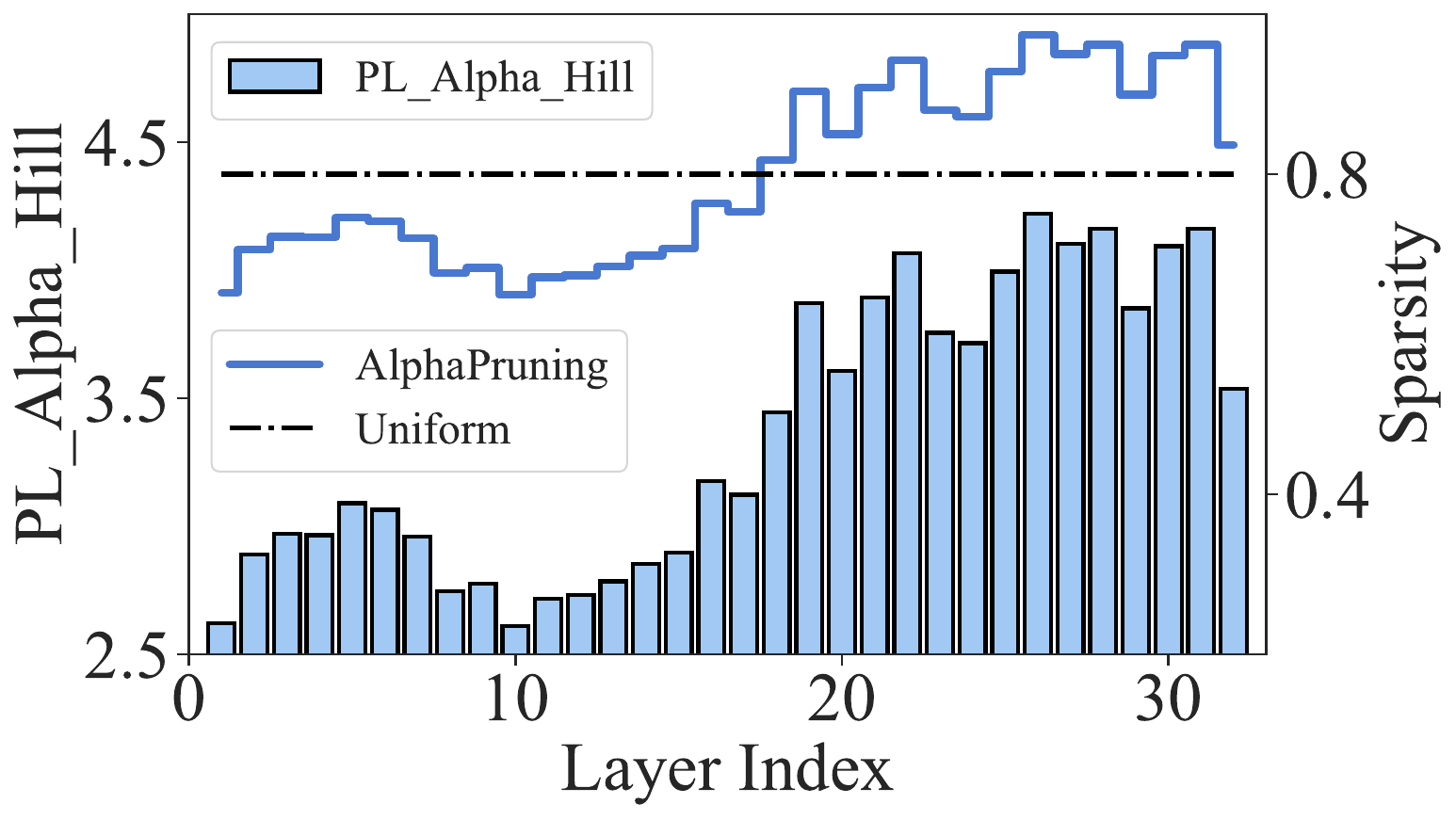}
\caption{{Comparing layerwise sparsities of \ourmethod and uniform sparsities, at 80\% global sparsity on LLaMA-7B.} The curves represent the layerwise sparsities, which are determined by \ALPHAHILL values shown by the histograms.
}
\vspace{-4mm}
\label{fig:distribution}
\end{wrapfigure}

\paragraph{{\bf Analyzing the layerwise sparsity distribution.}}
We elaborate on the layerwise sparsities assigned by \ourmethod and heavy-tailed metric \ALPHAHILL.
The metric and sparsities results of LLaMA-7B are presented in Figure~\ref{fig:distribution}.
We can observe that, the blue bar histograms demonstrate that different layers of one model show diverse \ALPHAHILL values, this indicates that these layers are not equally well-trained.
This conclusion is based on prior research on heavy-tails in weight matrices~\cite{martin2020predicting_NatComm, martin2019traditional}, this metric measures the heavy-tailed structure within the correlation matrix. 
This measurement indicates the amount of correlation among the weight matrix elements, with strong correlations leading to a more heavy-tailed empirical spectral density. 
Such a structure is often seen as a result~\cite{wang2024spectral} of extracting various useful correlations (or features) from data during optimization.

Consequently, smaller \ALPHAHILL layers (more heavy-tailed) contain a greater number of learned correlations. 
The larger \ALPHAHILL layers (less heavy-tailed) tend to retain fewer learned correlations, remaining closer to the random initialization state.
Figure~\ref{fig:distribution} shows that \ourmethod suggests that large \ALPHAHILL layers with fewer learned correlations should be allocated with larger sparsity, or being pruned more, while those small \ALPHAHILL layers should be allocated with lower sparsity.

\begin{figure}[!h]
    \centering
    \begin{subfigure}{0.6\linewidth}
    \includegraphics[width=\linewidth]{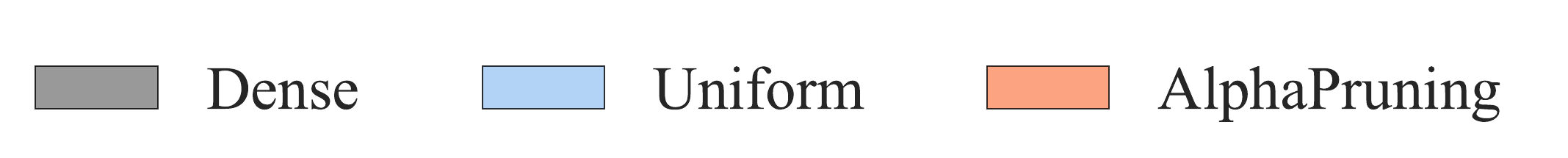}
    \end{subfigure} \\
    \begin{subfigure}{0.45\linewidth}
    \includegraphics[width=\linewidth]{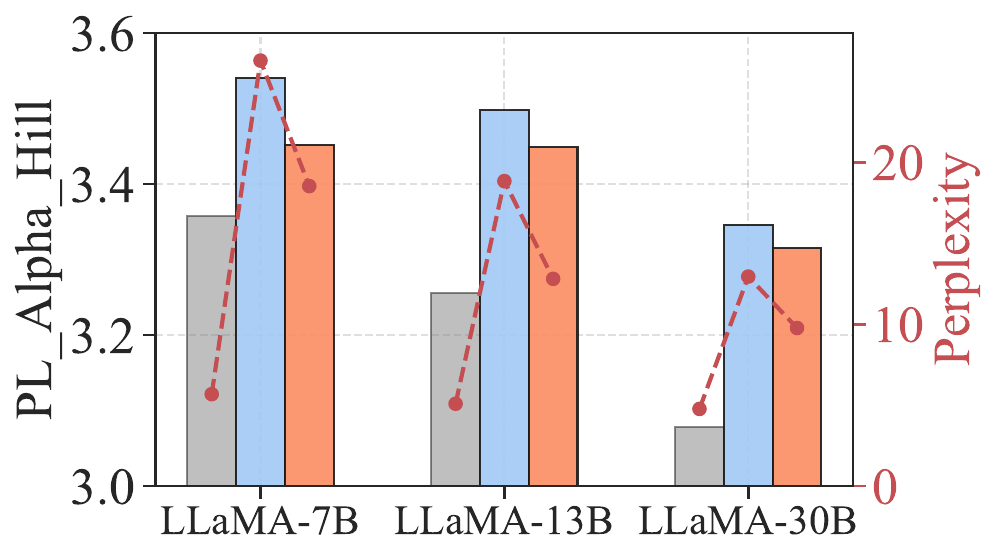}
    \caption{Model-wise heavy-tail analysis} \label{fig:ht-analysis-model}
    \end{subfigure}
    \begin{subfigure}{0.50\linewidth}
    \includegraphics[width=\linewidth]{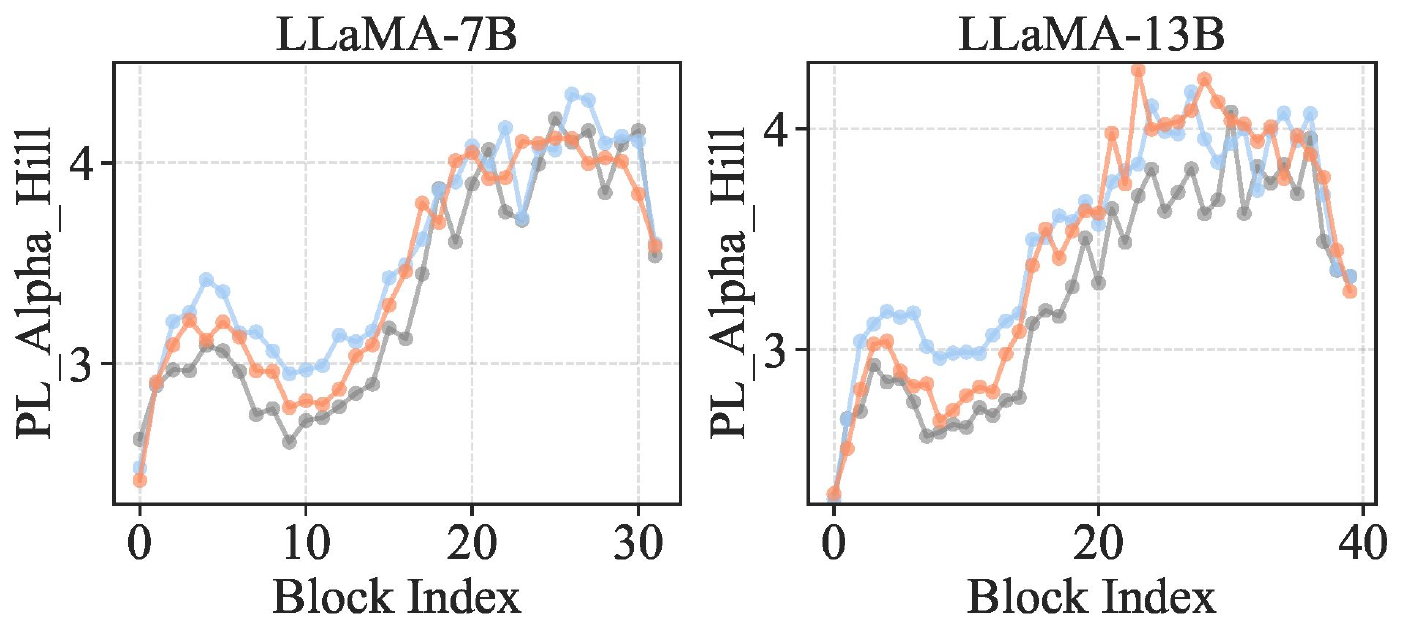}
    \caption{Layer-wise heavy-tail analysis}\label{fig:ht-analysis-layer}
    \end{subfigure}
    \caption{Analyzing the heavy-tail metric \ALPHAHILL (lower the better by HT-SR theory) and performance metric WikiText validation perplexity (lower the better) before and after pruning by baseline uniform pruning and \ourmethod. 
    (a) The metric value is reported by averaging over all layers within each model. The dashed lines represent the perplexity and the histograms represent the \ALPHAHILL value.
    (b) The metric is reported by averaging all the matrices within each LLM layer. \looseness-1
    }  \label{fig:ht-analysis}
\end{figure}

\paragraph{{\bf Analyzing \ALPHAHILL affected by pruning.}} 
We investigate how \ALPHAHILL values change before and after pruning.
According to HT-SR Theory~\cite{yang2023test, martin2020predicting_NatComm}, models or layers of higher quality typically exhibit lower \ALPHAHILL values. 
As observed in Figure~\ref{fig:ht-analysis-model}, dense LLaMA models (depicted in gray) consistently show a lower mean \ALPHAHILL, with larger models demonstrating an even lower mean. Furthermore, we can see both pruning methods lead to increments of the metric value, as pruning is often seen as damaging the quality of the model, the changes are well correlated with the perplexity.
More interestingly, \ourmethod not only outperforms the Uniform pruning in perplexity, but it also leads to a smaller mean of \ALPHAHILL. 
Figure~\ref{fig:ht-analysis-layer} delves deeper into this phenomenon by visualizing the metric value in a layer-wise manner.
It is noticeable that \ourmethod leads to lower \ALPHAHILL than the uniform pruning among the first several blocks.
This is due to the mechanism (Figure \ref{fig:distribution}) by which our method prunes the model based on the layer-wise \ALPHAHILL, and prunes less on these more heavy-tailed layers.

Based on these results, we can conclude that: 
(1) pruning a model to a larger sparsity generally hurts the model's task performance (e.g., perplexity), and this coincides with decreased model quality, as measured by \ALPHAHILL; and
(2) a better pruning method such as \ourmethod can obtain a sparse model with a smaller mean of layer-wise \ALPHAHILL, and this is achieved by pruning less aggressively in the dense layers with lower \ALPHAHILL layers.

\paragraph{Connections with Low-rank Approximation.} 
A commonly used compression technique that also involves analyzing the eigenspectrum of the weights is Low-Rank Approximation (LRA)~\cite{zhang2015accelerating,wen2017coordinating,xu2019trained,barsbey2021heavy}. 
We examine the apparent connection between our method, \ourmethod, and LRA from two perspectives. 
First, we study the relationship between the ESD used in our method and the low-rank properties often used in LRA, where we use \ALPHAHILL to measure HT and \STABLERANK to measure low-rank properties.
Second, we study the differing layer-wise assignment strategies adopted by the two methods.
Finally, we discuss how our findings relate to those of \citet{barsbey2021heavy}, a paper closely related to ours, showing that the results from both studies complement each other, offering different yet compatible insights.\looseness-1

\begin{figure}[!th]
    \centering
    \begin{subfigure}[t]{0.30\linewidth}
    \includegraphics[width=\linewidth,keepaspectratio]{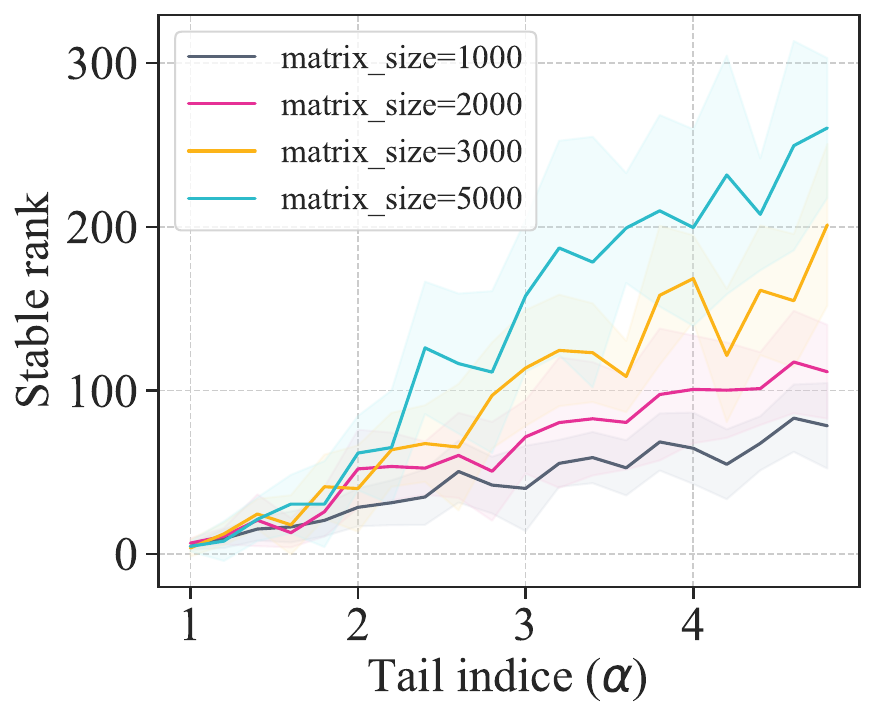}
    \caption{\STABLERANK of ESD sampled from Pareto distribution}\label{fig:LRA_results-pare}
    \end{subfigure} 
    \centering
    \begin{subfigure}[t]{0.33\linewidth}
    \includegraphics[width=\linewidth,keepaspectratio]{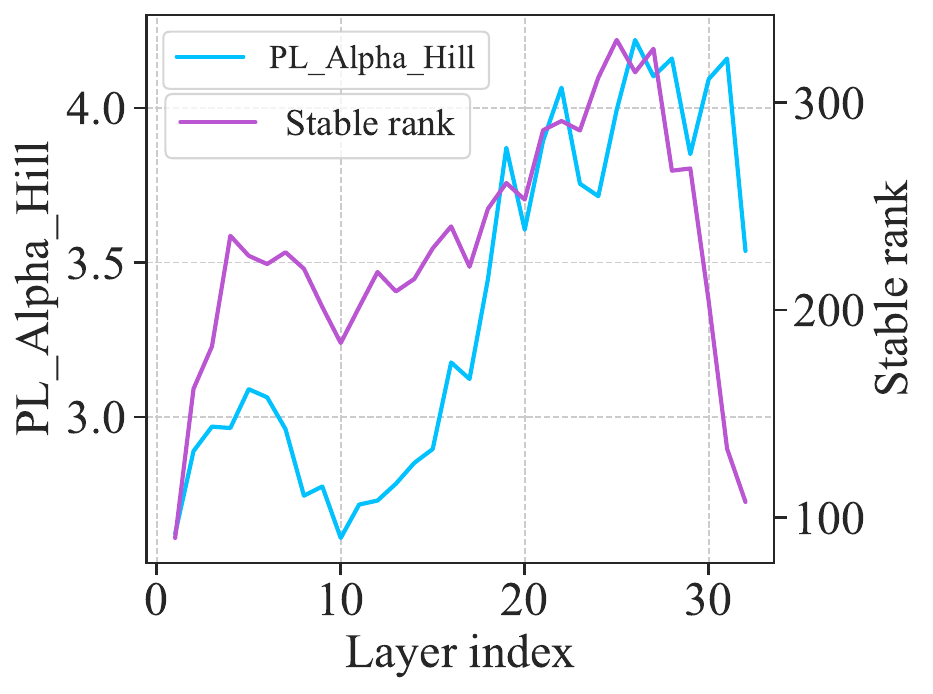}
    \caption{Layerwise metric values of LLaMA-7B}\label{fig:LRA_results-model}
    \end{subfigure}
    \centering
    \begin{subfigure}[t]{0.31\linewidth}
   \includegraphics[width=\linewidth,keepaspectratio]{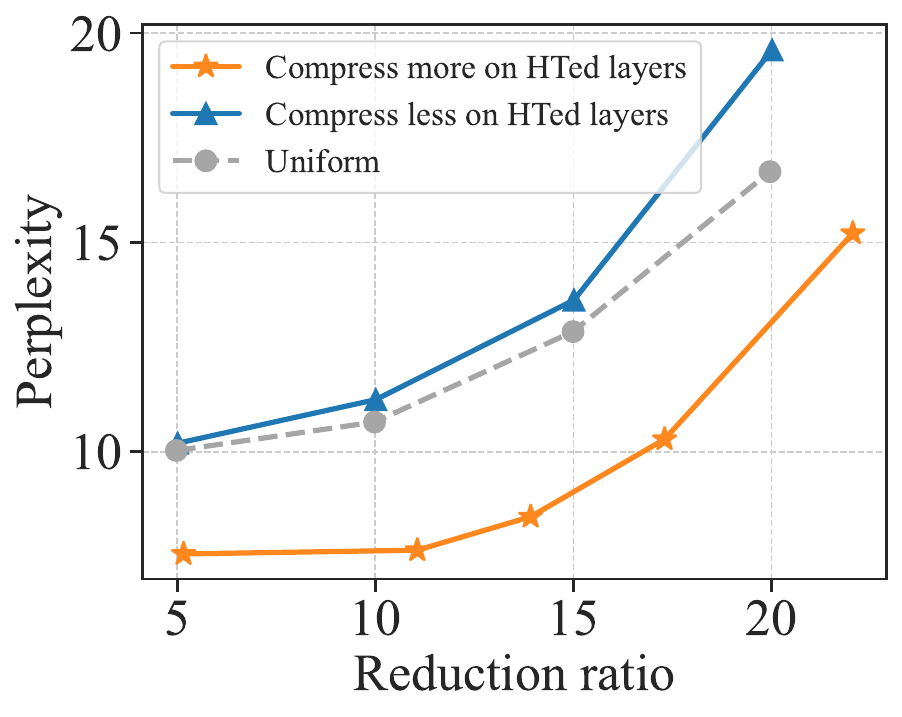}
   \caption{Comparing layer-wise assignment strategies for LRA}\label{fig:LRA_results-method}
   \end{subfigure} 
\caption{\textbf{Analyzing ESD properties and assignment strategies for LRA.} (a) \STABLERANK and \ALPHAHILL show a similar pattern (the more heavy-tailed, the more low-ranked) across different ESDs sampled from Pareto distribution. (b) The layer-wise \ALPHAHILL and \STABLERANK of the LLaMA-7B model exhibit a similar trend. (c) Comparing two assignment strategies for LRA, and ``Compress more on HTed layers'' is better. This finding is opposite to pruning-based methods, which find that ``Compress less on HTed layers'' is better.
}~\label{fig:LRA_results}
\end{figure}

For the first point, we find a strong positive correlation between the HTed properties and the low-rank properties used in LRA: as the ESD becomes more HTed, it also becomes more low-ranked. To demonstrate this, we sampled eigenvalues from an IID Pareto distribution with varying tail indices to form ESDs and then measured \STABLERANK on these ESDs. A lower \ALPHAHILL indicates a more HTed distribution, while a lower \STABLERANK indicates a more low-ranked structure. Figure~\ref{fig:LRA_results-pare} shows that \STABLERANK and \ALPHAHILL are positively correlated across different matrix sizes, suggesting a relationship between HTed and low-ranked properties. In Figure~\ref{fig:LRA_results-model}, we further verify this by measuring \ALPHAHILL and \STABLERANK across different layers of the pre-trained LLaMA-7B model. The results reveal that both metrics follow a similar trend: shallow layers show lower values, while deeper layers exhibit higher values, indicating consistent behavior across layers.
Therefore, we can see that the key metrics used in \ourmethod and LRA are highly correlated.

For the second point, we clarify that the strategies for assigning higher or lower compression to HTed or low-ranked layers differ between \ourmethod and LRA. \ourmethod assigns lower compression ratios (less sparsity) to HTed layers.
In contrast, LRA assigns higher compression to low-ranked layers~\cite{zhang2015accelerating,wen2017coordinating,xu2019trained}, which, as established earlier, tend to be HTed. Therefore, the two methods seemingly use different procedures in assigning layerwise compression ratios.
In Figure~\ref{fig:LRA_results-method}, we empirically verify that, for LRA, assigning higher compression to HTed layers is indeed more beneficial than its opposite for LLMs, as shown by the results of applying both to LLaMA-7B models. This finding is opposite to the assignment method used in \ourmethod.
We hypothesize that these differences arise from the distinct mechanisms and principles of each method. 
In more detail, \ourmethod, which is pruning-based, removes elements from the weight matrices, affecting the entire eigenspectrum. LRA, however, removes only the smallest eigenvalues, leaving the larger eigenvalues intact.
The guiding principle of \ourmethod is also different from LRA. It aims to make the model more deterministic and less random by preserving weights corresponding to well-trained layers. It does this by preserving HTed layers that contain more signals and removing light-tailed layers that, according to HT-SR theory, resemble randomly initialized weight matrices.
In some sense, it is similar to how decision trees choose branches to reduce entropy.
LRA, on the other hand, focuses on applying more compression on low-rank matrices where the largest eigenvalues dominate. This allows for minimal impact on reconstruction loss when removing small eigenvalues. 

Lastly, \citet{barsbey2021heavy} concluded that models with lower HT measures are generally more compressible than other models, focusing on cross-model comparisons.
In contrast, our work examines compressibility within a single model and suggests that, within a model, layers with lower HT values are less compressible than other layers.
Our insight supports layer-wise pruning strategies, such as \ourmethod, which applies less compression to more HTed layers. 
We empirically confirm that both conclusions are complementary.
Following the setup of \citet{barsbey2021heavy}, detailed in Appendix~\ref{app:barsbey}, we reproduced their findings in Figure~\ref{fig:prunable-vary-modelalpha} and demonstrated compatibility with \ourmethod's assignment strategies.
We generated four trained models with varying model-wise HT measures and pruned them using different layer-wise strategies. 
The results verified that models with lower HT metrics are more compressible. 
In Figure~\ref{fig:prunable-vary-layeralpha}, we further validated our findings using the same models.
Among four models with varying model-wise HT values, we confirmed that HTed layers are less compressible and that pruning HTed layers less (which \ourmethod does) is more effective.

\section{Conclusion}
We have used methods from HT-SR Theory to develop improved methods for pruning LLMs.  
The basic idea is to analyze the ESDs of trained weight matrices and to use shape metrics from these ESDs to measure how much to prune a given layer, with less well-trained layers, as measured by these shape metrics from HT-SR Theory, being pruned more aggressively.
Our extensive empirical evaluation demonstrates that \ourmethod offers a straightforward yet effective way of determining the layer-wise sparsity ratios. 
Our analysis reveals that different layers of an LLM are not equally trained (typically, the ESDs of early layers are more HT and thus are more well-trained, compared to later layers), and that shape-based ESD metrics work better for layer quality prediction in pruning pipelines than scale-based ESD metrics. 
\ourmethod achieves higher sparsity, without severely hurting performance, and also smaller values of \ALPHAHILL after pruning.
\ourmethod is also compatible with multiple existing LLM pruning methods and is expected to be integrated with future ones, as long as the methods allow specifying layerwise ratios.
\section*{Acknowledgements.}
We want to thank Alex Zhao, Elicie Ye, Zhuang Liu, Xiangyu Yue, and Tianyu Pang for their helpful discussions. 
Michael W. Mahoney would like to acknowledge the UC Berkeley CLTC, ARO, IARPA (contract W911NF20C0035), NSF, and ONR for providing partial support of this work. 
Yaoqing Yang would like to acknowledge support from DOE under Award Number DE-SC0025584, DARPA under Agreement number HR00112490441, and Dartmouth College.
Our conclusions do not necessarily reflect the position or the policy of our sponsors, and no official endorsement should be~inferred.

\bibliographystyle{plainnat}
\bibliography{example_paper}

\newpage
\appendix
\clearpage
\section*{Appendix}

\section{Impact Statements}\label{sec:impact-statement}
This paper leverages HT-SR Theory to design improved layerwise pruning ratios for LLMs.
Although the proposed method could be applied to compress models with adverse applications, we do not see any
immediate negative societal impacts stemming from the algorithm itself. 
Indeed, we see a lot of societal value in proposing our method to the community.
Through the implementation of effective layerwise sparsity, we can achieve substantial reductions in the parameters of LLMs while retaining their functionality. 
Consequently, this advancement facilitates the deployment of LLMs in resource-constrained devices, accelerates the predictions for resource-limited LLM services, and contributes to the sustainability of LLM technologies.

\section{Limitation}\label{sec:limitation}
This paper's empirical evaluation focuses on post-training pruning methods without fine-tuning, in alignment with recent research in LLM pruning, such as OWL, Wanda, and SparseGPT. 
This is due to the substantial computational resources required to restore heavily pruned LLMs to their original performance. 
Nonetheless, our experiments with limited fine-tuning have demonstrated that the proposed pruning method can achieve only a mild performance drop while yielding significant efficiency improvements.

\ifisTR\else

\fi

\section{Definitions of HT-SR metrics}
\label{sec:supp-metrics}

Here, we define the shape metrics, beyond \ALPHAHILL, that we use in our analysis.
\begin{itemize}[noitemsep,topsep=0pt]
    \item (\ALPHAHAT) The \ALPHAHAT metric~\cite{martin2020predicting_NatComm} has been shown to be effective at predicting generalization. It is the variant of PL exponent $\alpha$ (\ALPHA) weighted by the log maximum eigenvalue $\log\lambda^\text{max}$(\LOGSPECTRALNORM):
\begin{equation}
\label{eqn:alphahat}
    \ALPHAHAT = \alpha \log \lambda ^\text{max}.
\end{equation}
    \item (\STABLERANK) The \STABLERANK metric is a norm-adjusted measure of the scale of the ESD, and previous work~\cite{yang2023test} has shown its strong correlation with \ALPHA. For a weight matrix $\mathbf{W}$, it can be calculated as:
\begin{equation}
\label{eqn:stablerank}
    \STABLERANK = \left \| \mathbf{W} \right \| ^2_F / \left \| \mathbf{W} \right \| _2^2 .
\end{equation}
    \item (\ENTROPY) For a weight matrix $\mathbf{W}$, the \ENTROPY metric is defined as
\begin{equation}\label{eqn:entropy}
    \ENTROPY = \frac{-1}{\log_{}{R(\mathbf{W})}}\sum_{i}^{}  p_i\log_{}{p_i}.
\end{equation}
where $p_i=v_i^2 / \sum_{i}^{}v_i^2$, $v_i$ is the $i$-th singular value of $\mathbf{W}$, and $R(\mathbf{W})$ refers to the rank of $\mathbf{W}$.
\end{itemize}

\section{Further analysis of results}

\subsection{Post-pruning layer-wise heavy-tail analysis}\label{more-analysis}
\begin{figure}[!h]
    \centering
    \begin{subfigure}{0.4\linewidth}
    \includegraphics[width=\linewidth]{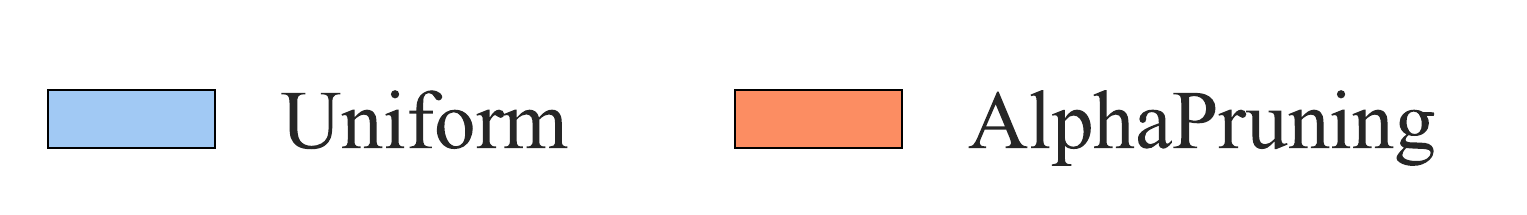}
    \end{subfigure} \\
    \hspace{0.02\linewidth}
    \begin{subfigure}{0.3\linewidth}
    \includegraphics[width=\linewidth]{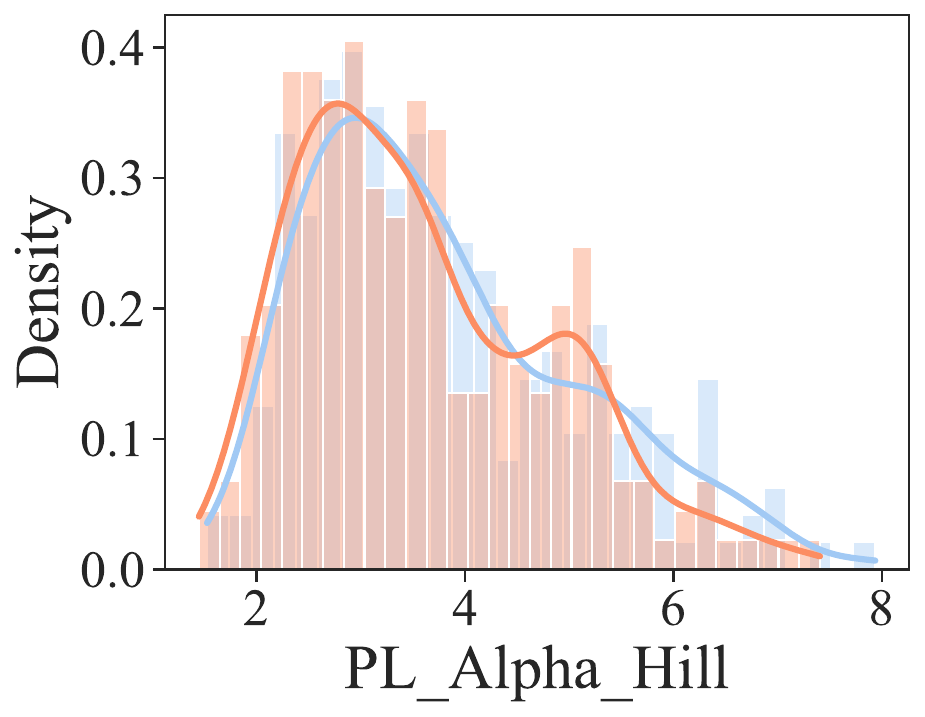}
    \caption{LLaMA-V3-8B} 
    \end{subfigure}
    \hspace{0.02\linewidth}
    \begin{subfigure}{0.3\linewidth}
    \includegraphics[width=\linewidth]{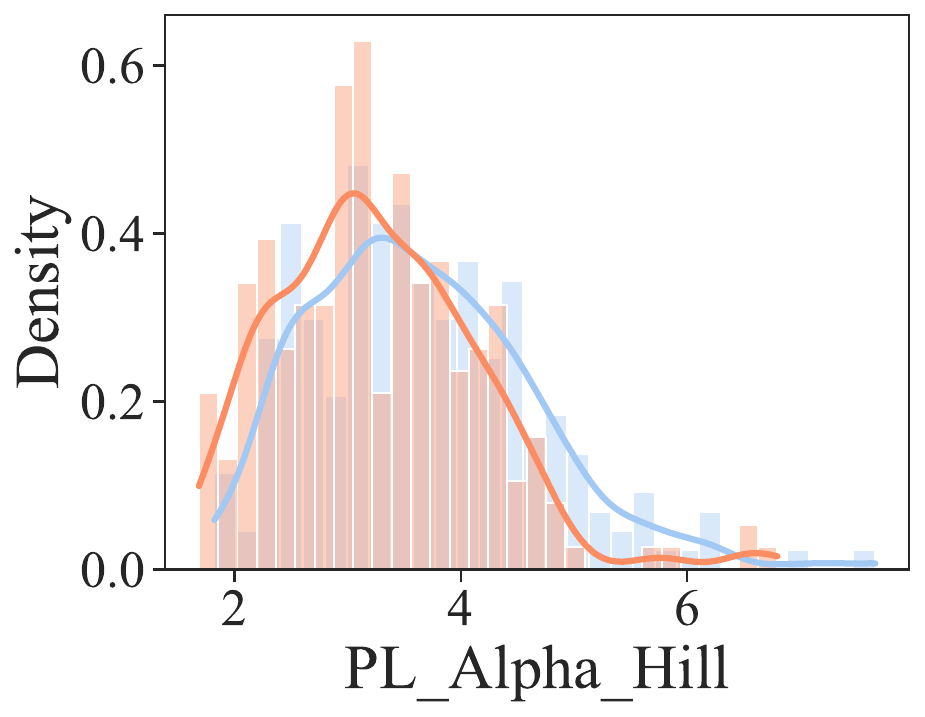}
    \caption{Vicuna-7B}
    \end{subfigure}
    \hspace{0.02\linewidth}
    \begin{subfigure}{0.3\linewidth}
    \includegraphics[width=\linewidth]{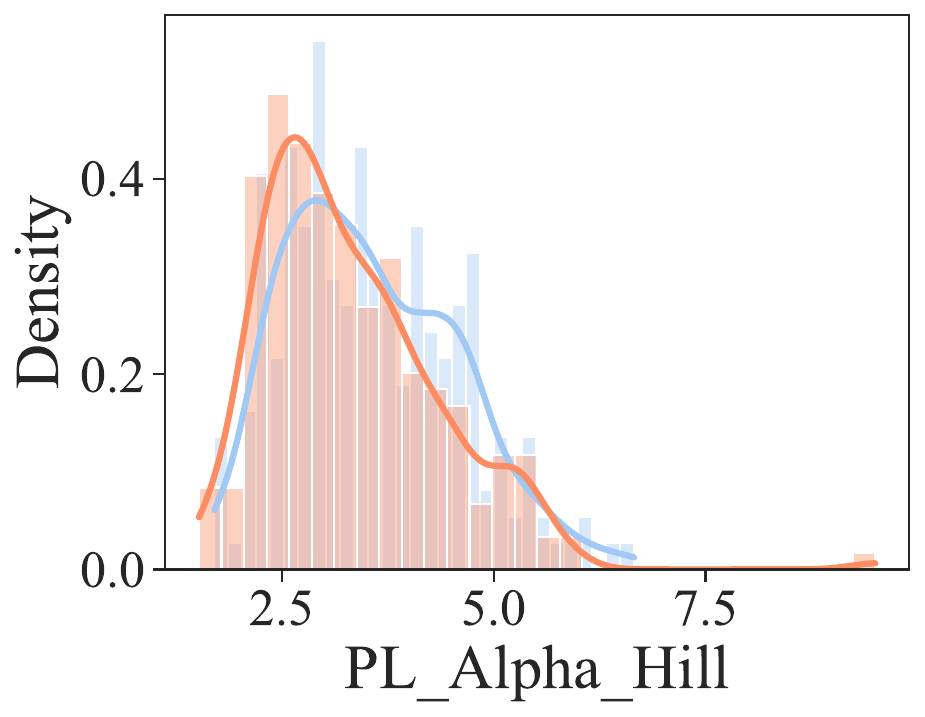}
    \caption{Mistral-7B}
    \end{subfigure} 
    \caption{Analyzing the layer-wise heavy-tail metric \ALPHAHILL (lower the better by HT-SR theory) after pruning by baseline uniform pruning and \ourmethod. 
    }  \label{fig:ht-analysis-layerwise}
\end{figure}

We investigate layer-wise \ALPHAHILL values after pruning by Uniform pruning and \ourmethod on more advanced LLMs (LLaMA-V3-8B, Vicuna-7B, Mistral-7B). According to HT-SR Theory, models or layers of higher
quality typically exhibit lower \ALPHAHILL values. As observed in Figure~\ref{fig:ht-analysis-layerwise}, \ourmethod leads to a smaller layer-wise \ALPHAHILL. This is due to the mechanism (\ref{fig:teaser}) by which our method prunes the model based on the layer-wise \ALPHAHILL, and prunes less on these more heavy-tailed layers.

\begin{figure}[!thb]
    \centering
    \includegraphics[width=0.8\linewidth]{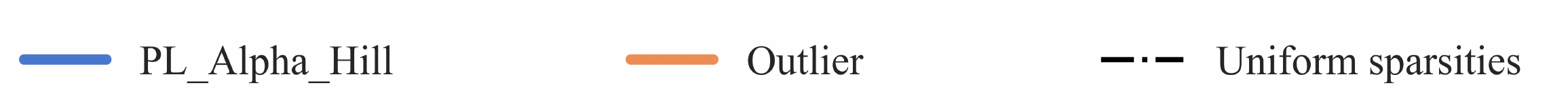} 
    \includegraphics[width=0.32\linewidth]{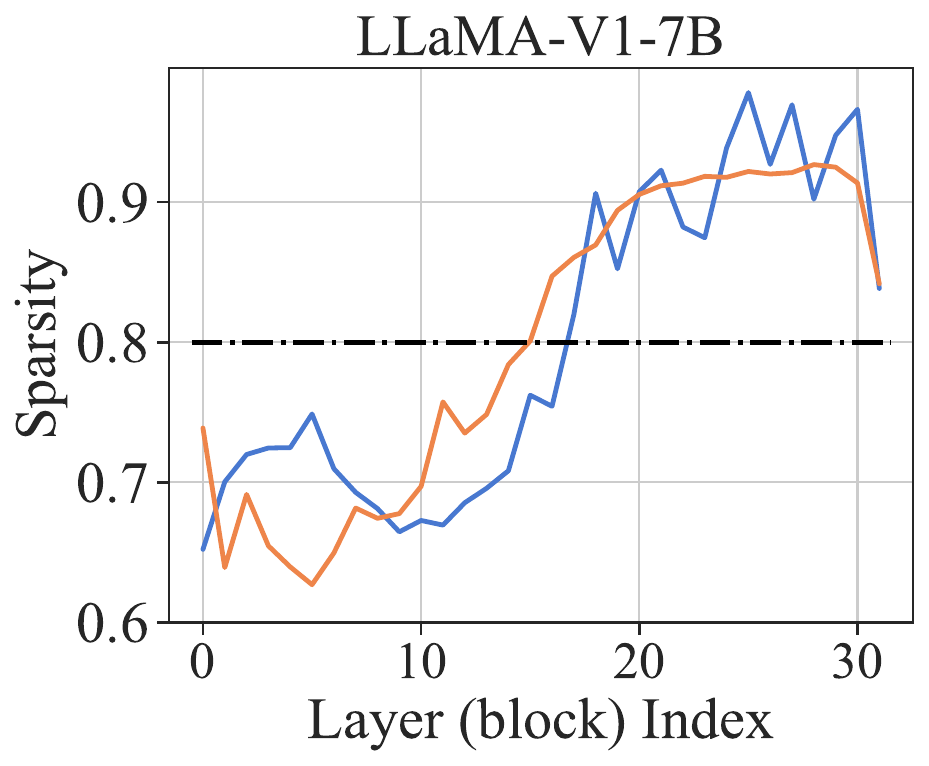}
    \includegraphics[width=0.32\linewidth]{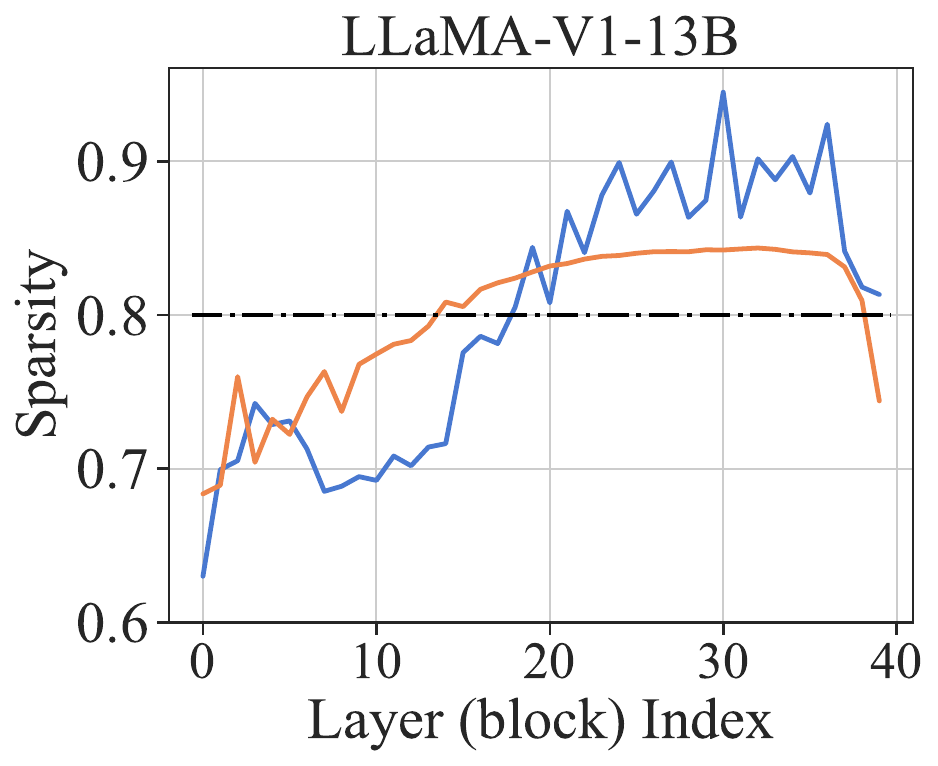}
    \includegraphics[width=0.32\linewidth]{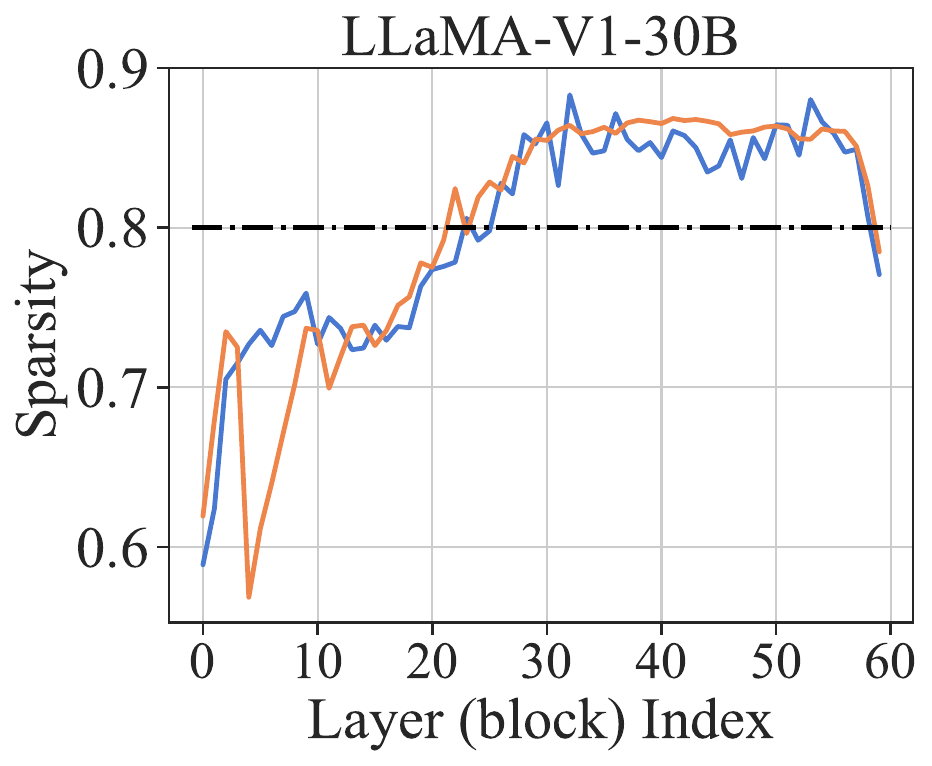}
    \caption{ \textbf{Comparing layer-wise sparsity distributions allocated by
AlphaPruning (blue, ours) and OWL (orange)}. While both methods
show similar overall trends, \ourmethod generates a more granular
distribution with distinct differences between consecutive layers.}\vspace{-4mm}
\label{fig:distribution_owl}
\end{figure}

\subsection{Comparison with other LLM layer quality metrics}
In addition to \ourmethod proposed in our work, \citet{gromov2024unreasonable, men2024shortgpt} are other studies that investigated methods that measure whether a layer is well-trained or not, demonstrating LLMs layers are not equally well-trained. \citet{gromov2024unreasonable} developed a method that assesses the similarity between the representations at different layers, defined as the angular distance between feature vectors. They found that deeper layers are more similar to their neighboring layers than shallow layers, suggesting that LLMs may not fully use the parameters in these deeper layers, indicating these layers are not well-trained. Similarly, \citet{men2024shortgpt} introduced a metric called Block Influence, which measures the impact of each transformer block on hidden states to gauge layer significance. Their findings showed varying degrees of ineffectiveness/redundancy across layers, suggesting that these layers are not well-trained.

Besides, in Figure~\ref{fig:distribution_owl}, we compare the sparsity allocation of \ourmethod with OWL. We show that the general trends of sparsity distribution generated by the two methods are similar, with lower sparsities allocated to earlier layers and higher sparsities allocated to deeper layers. However, our method produces a more granular distribution with clearer distinctions between consecutive deep layers, resulting in improved pruning performance.

\subsection{Comparison with conclusions of \citet{barsbey2021heavy}}
\label{app:barsbey}

\begin{figure}[!tbh]
\centering
\begin{subfigure}{0.35\linewidth}
\includegraphics[width=\linewidth]{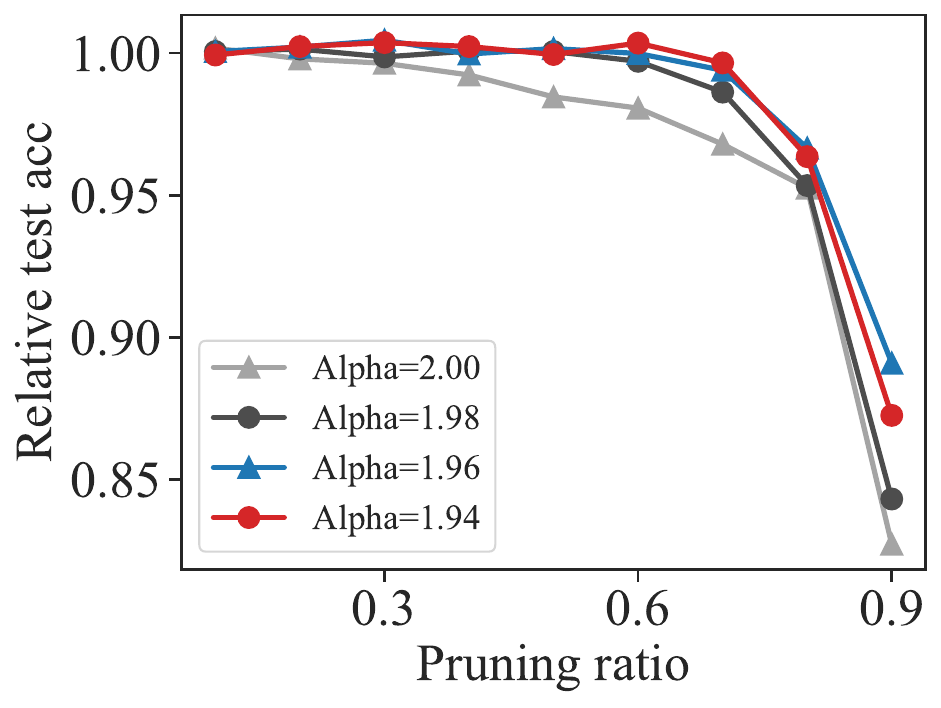}
\caption{Uniform Pruning}
\end{subfigure}
\hspace{10mm}
\begin{subfigure}{0.35\linewidth}
\includegraphics[width=\linewidth]{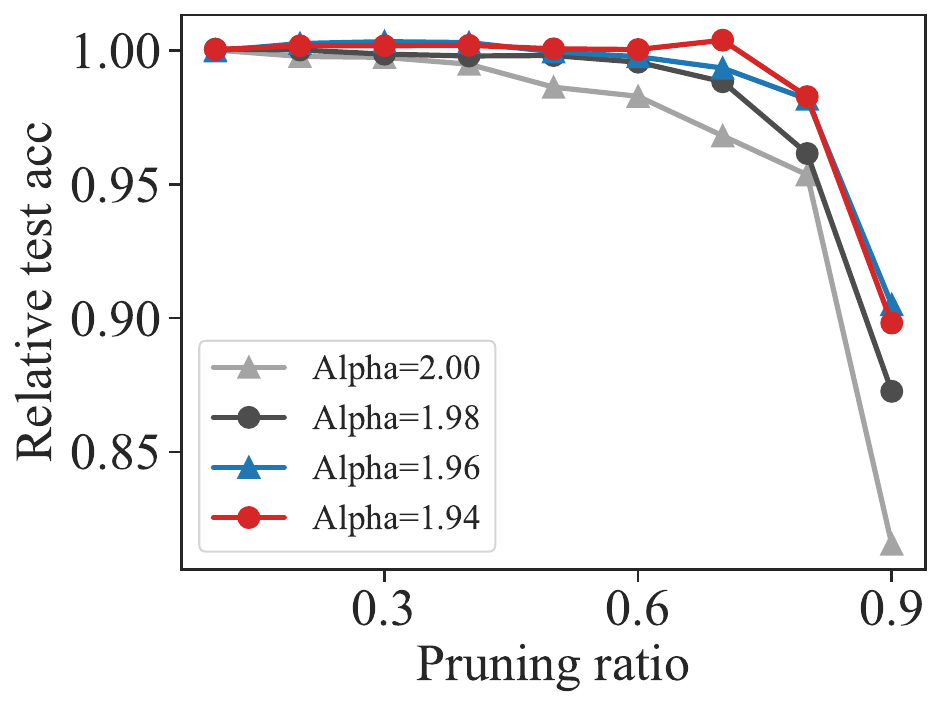}
\caption{\ourmethod}
\end{subfigure}
\caption{\textbf{Comparison of pruning models with varying model-wise HT measures (Alpha).} Models with higher HT measures are more prunable using both Uniform pruning and \ourmethod. Relative accuracy is calculated as the post-pruning accuracy divided by the pre-pruning accuracy. The experiments used FCNs trained on CIFAR-10.}\label{fig:prunable-vary-modelalpha}
\end{figure}

\begin{figure}[!tbh]
\centering
\begin{subfigure}{0.85\linewidth}
\includegraphics[width=\linewidth]{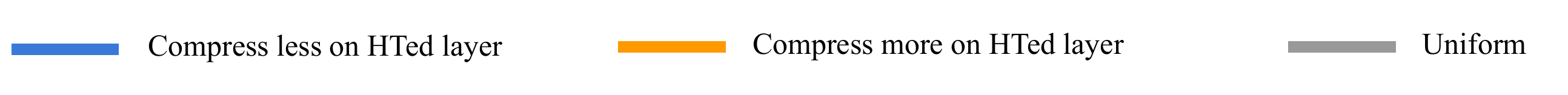}
\end{subfigure} \\
\begin{subfigure}{0.24\linewidth}
\includegraphics[width=\linewidth]{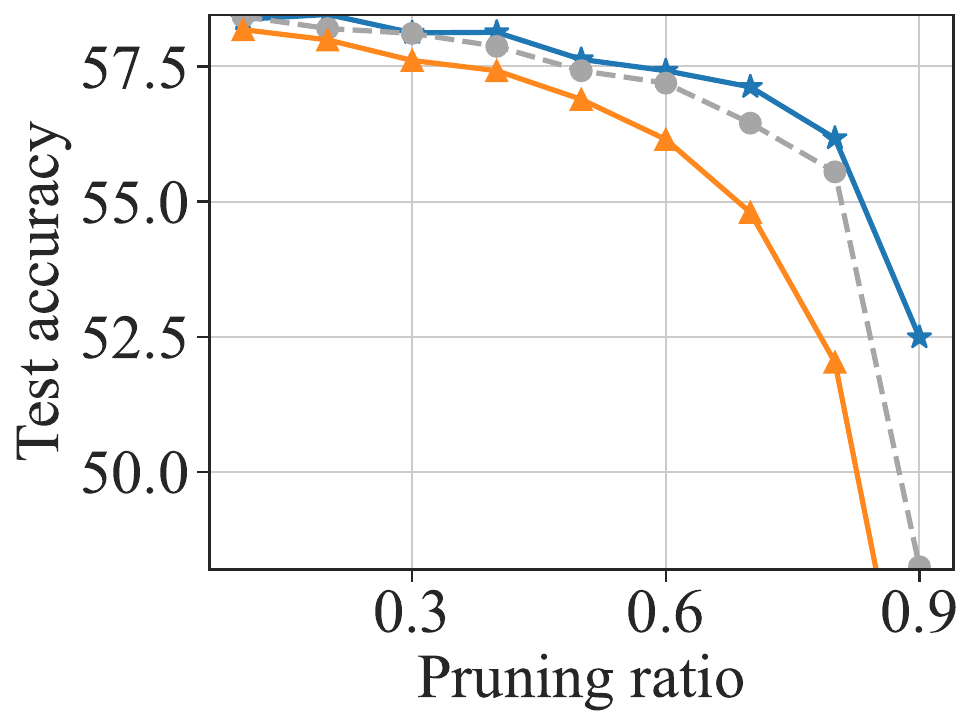}
\caption{Alpha=2.00}
\end{subfigure}
\begin{subfigure}{0.24\linewidth}
\includegraphics[width=\linewidth]{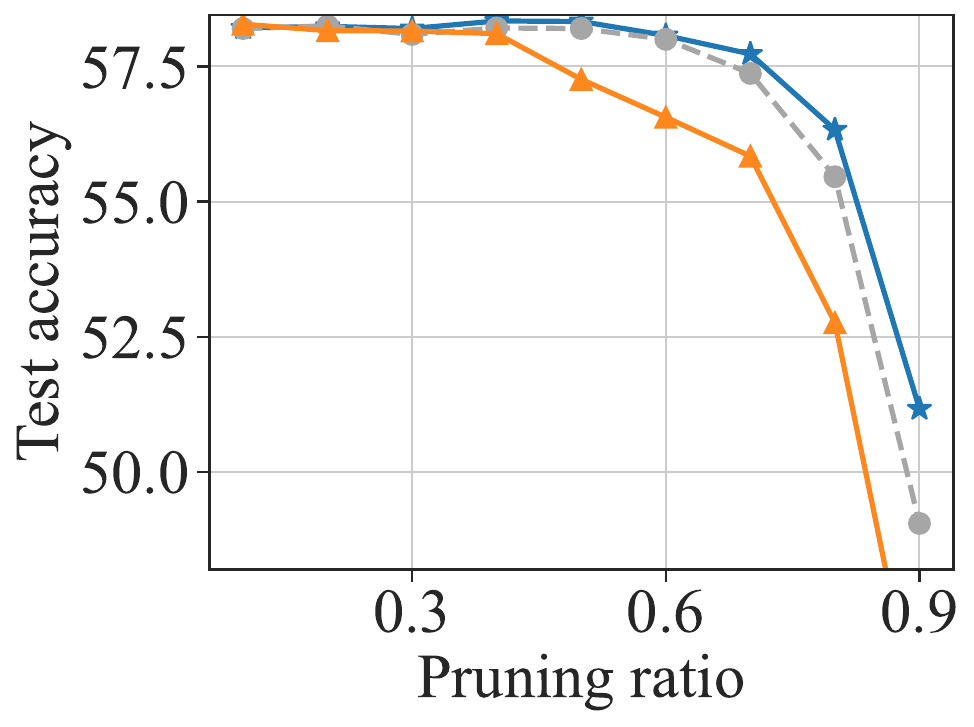}
\caption{Alpha=1.98}
\end{subfigure}
\begin{subfigure}{0.24\linewidth}
\includegraphics[width=\linewidth]{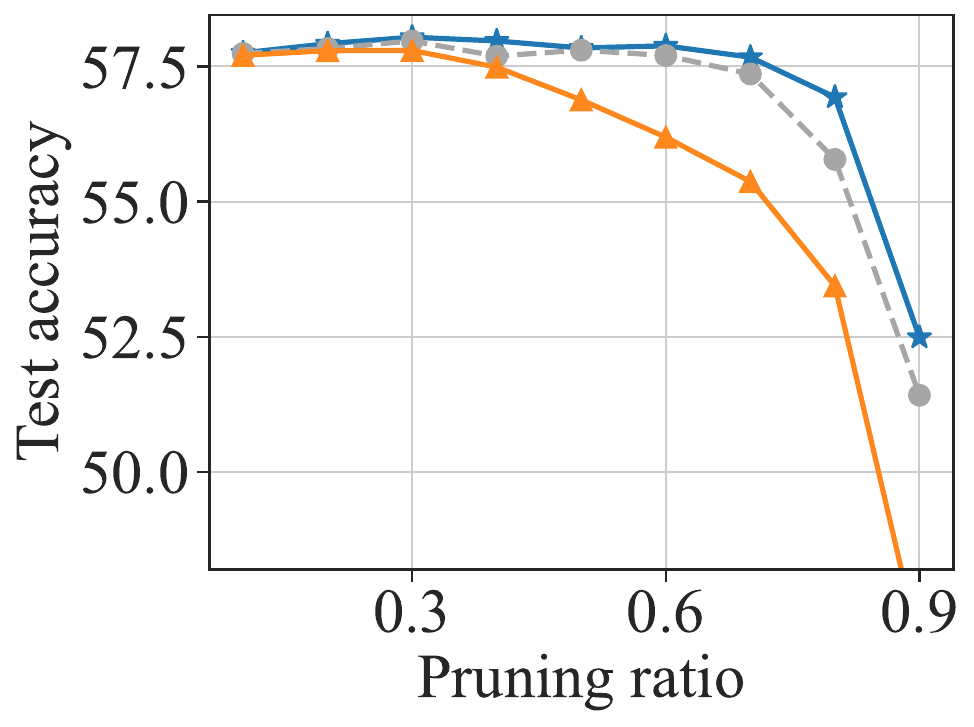}
\caption{Alpha=1.96}
\end{subfigure}
\begin{subfigure}{0.24\linewidth}
\includegraphics[width=\linewidth]{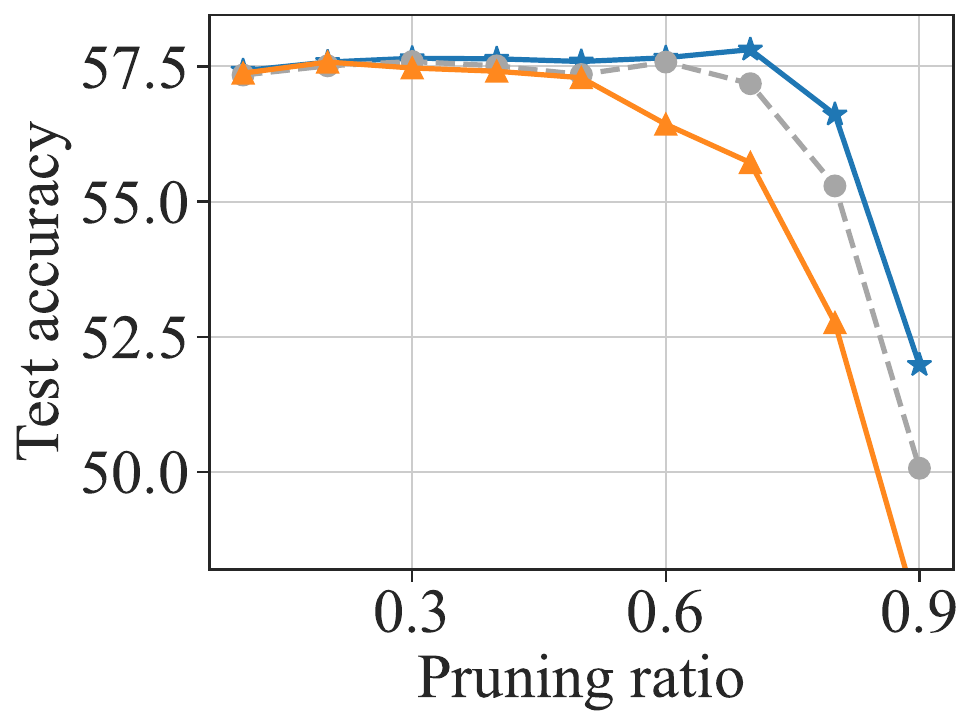}
\caption{Alpha=1.94}
\end{subfigure}
\caption{\textbf{Comparison of our proposed layer-wise assignment strategy with other strategies on models with varying model-wise HT measures.} ``Compress less on HTed layer'' (\ourmethod) consistently outperforms Uniform pruning across different models, while ``Compress more on HTed layer'' leads to worse performance compared to Uniform pruning. The experiments used FCNs trained on CIFAR-10.\looseness-1}\label{fig:prunable-vary-layeralpha}
\end{figure}

\citet{barsbey2021heavy} studied a problem related to ours, and they showed that models with more HTed weights had improved compressibility.
We adopt the experimental setup of \citet{barsbey2021heavy}, training fully connected networks (FCNs) with six hidden layers on the CIFAR-10 dataset. The HT measure used in their study referred to as Alpha, quantifies the HTed structure of the weight parameters. 
The model-wise HT measure is influenced by the ``temperature'', defined as the ratio of learning rate to batch size. Higher temperatures yield models with lower HT measures.
Our first experiment reproduces the finding of \citet{barsbey2021heavy}, comparing the compressibility of models with varying model-wise HT measures. 
It also tries to verify if the conclusions hold when using our proposed layer-wise strategies.
The results shown in Figure~\ref{fig:prunable-vary-modelalpha} support our hypotheses.
The second experiment compares the effectiveness of these layer-wise strategies and also checks if the findings are consistent across models with different model-wise HT measures. 
The results shown in Figure \ref{fig:prunable-vary-layeralpha} again confirm the hypothesis.

\section{Ablation study of sparsity allocation}\label{app:ablation-allocation}

In Section~\ref{sec:allocation}, we introduced the range hyperparameters $s_1$ and $s_2$ to control the non-uniformity of layer-wise sparsities. To simplify, we define $\tau$ such that $s_1 = 1-\tau, s_2 = 1+\tau$.
\ourmethod allocates sparsity on a ``per-block'' basis, where all matrices within a block receive the same sparsity, determined by averaging the \ALPHAHILL values across matrices within that block.
Alternatively, sparsity can be allocated on a ``per-matrix'' basis, allowing different sparsities for individual matrices based on their \ALPHAHILL values.
The ablation study on comparing per-matrix and per-block choices is presented in \ref{sec:app-per-matrix-block}. The ablation study on comparing different mapping functions is shown in \ref{app:diff-map-func}.

\begin{figure}[!thb] 
    \centering
    \includegraphics[width=0.5\linewidth,keepaspectratio]{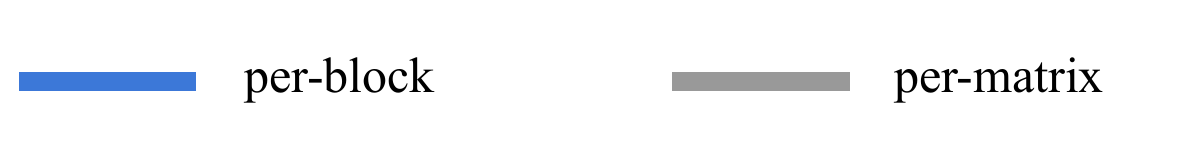} 
    \\
    \includegraphics[width=0.24\linewidth,keepaspectratio]{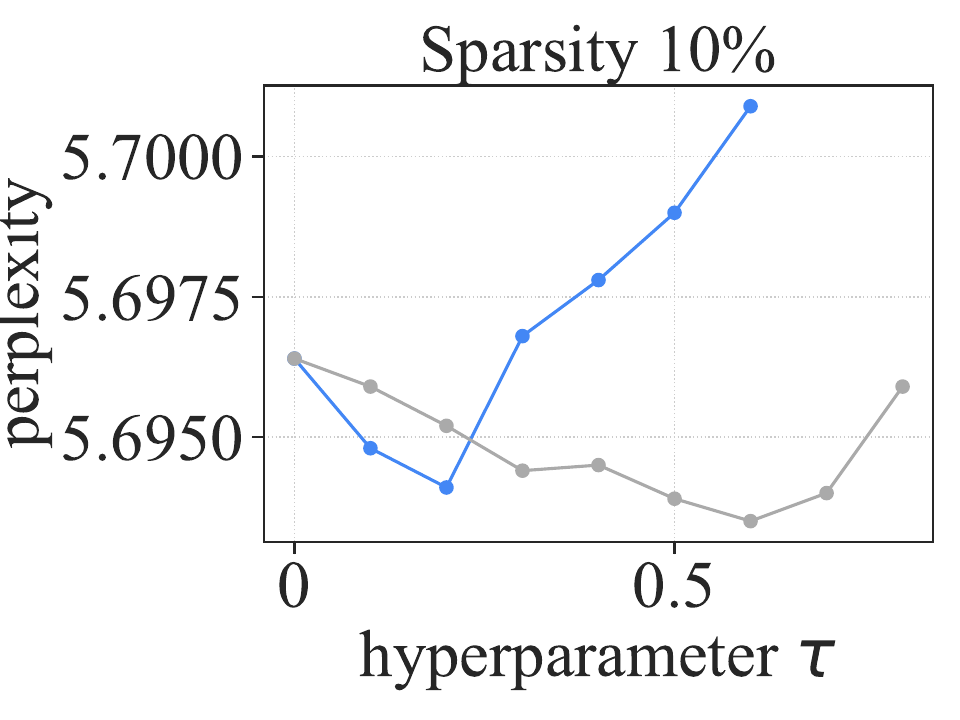}
    \includegraphics[width=0.24\linewidth,keepaspectratio]{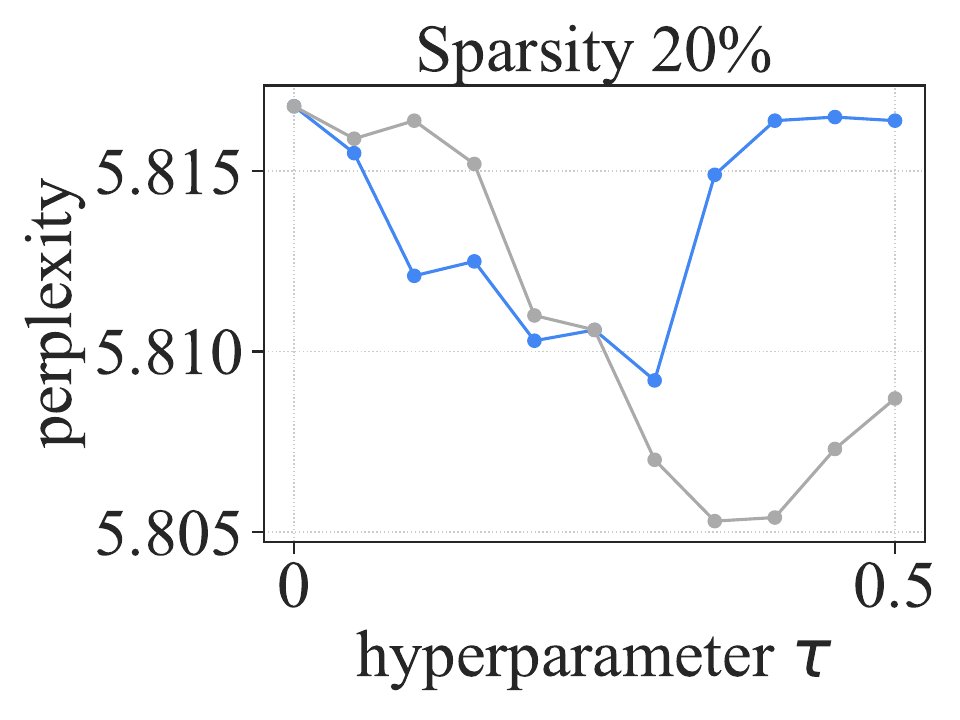}
    \includegraphics[width=0.24\linewidth,keepaspectratio]{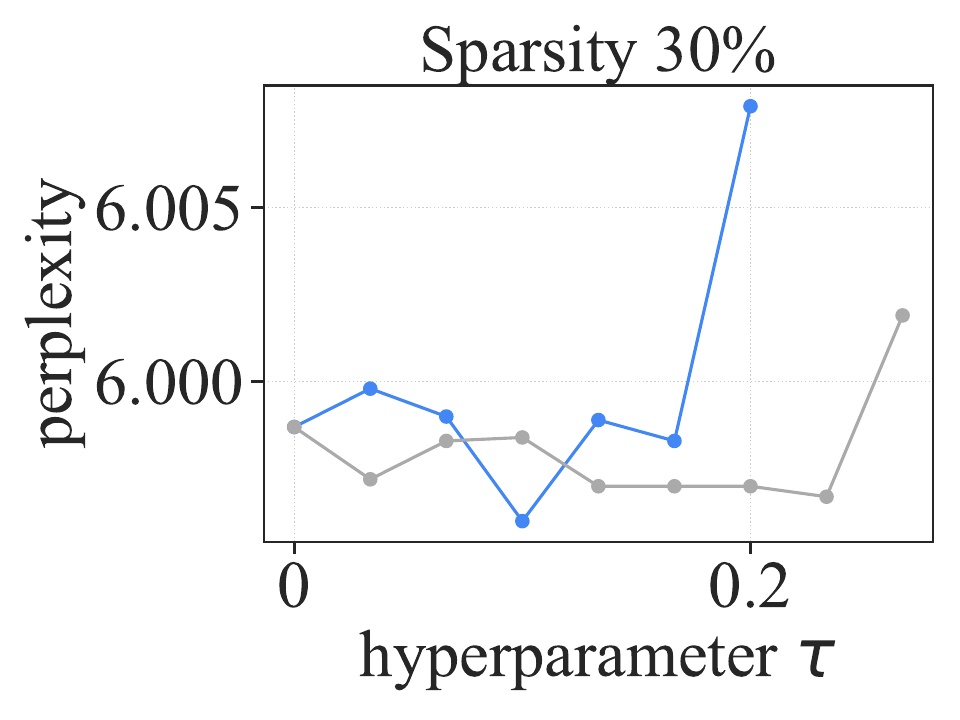}
    \includegraphics[width=0.24\linewidth,keepaspectratio]{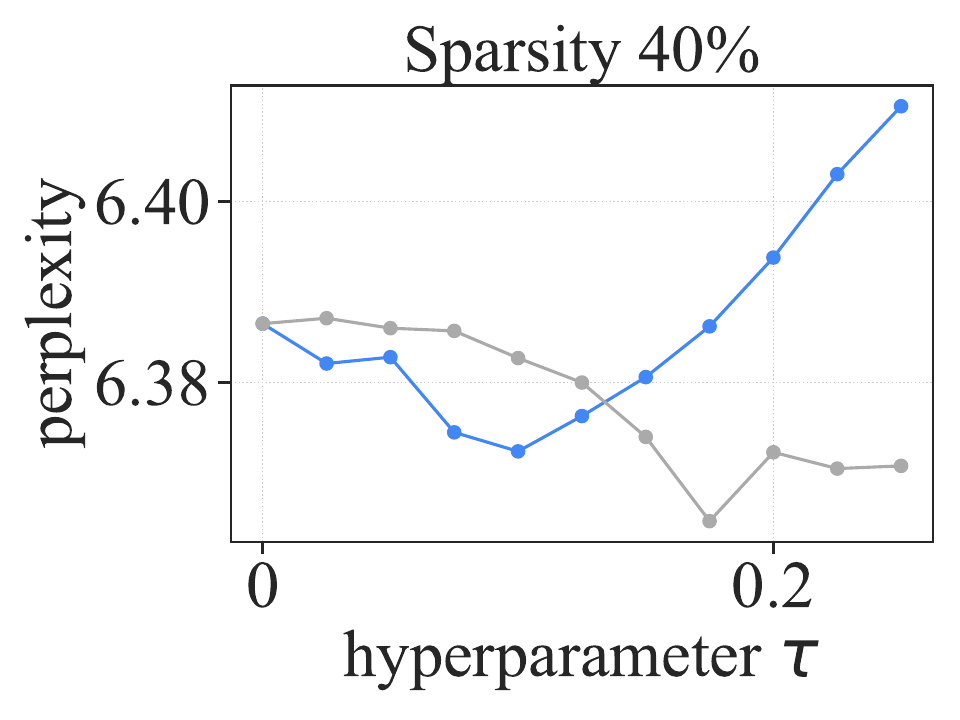}

    \includegraphics[width=0.24\linewidth,keepaspectratio]{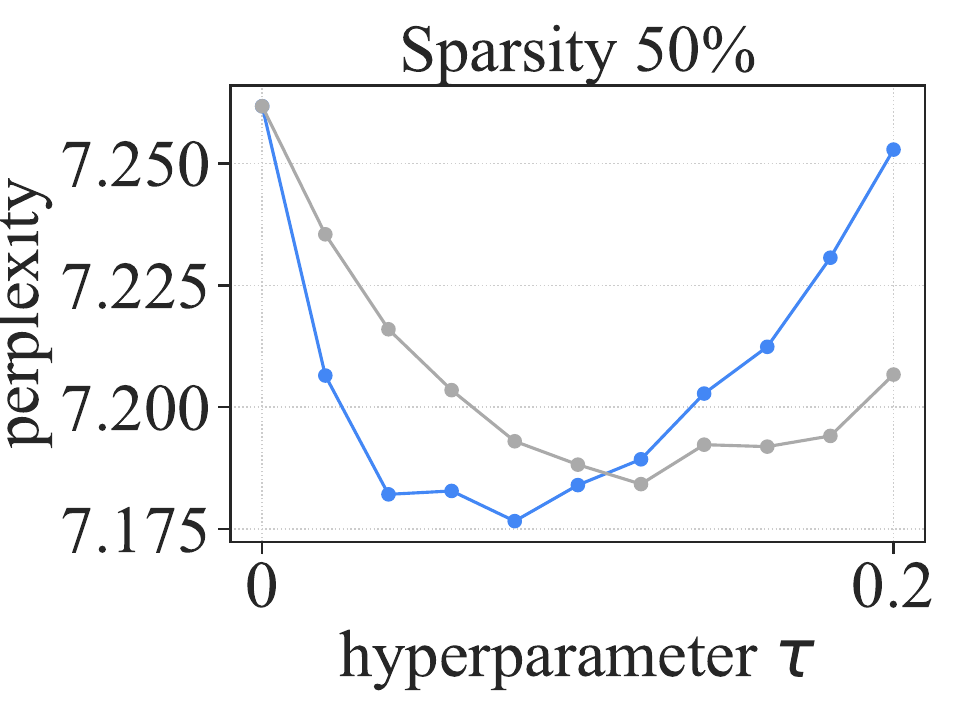}
    \includegraphics[width=0.24\linewidth,keepaspectratio]{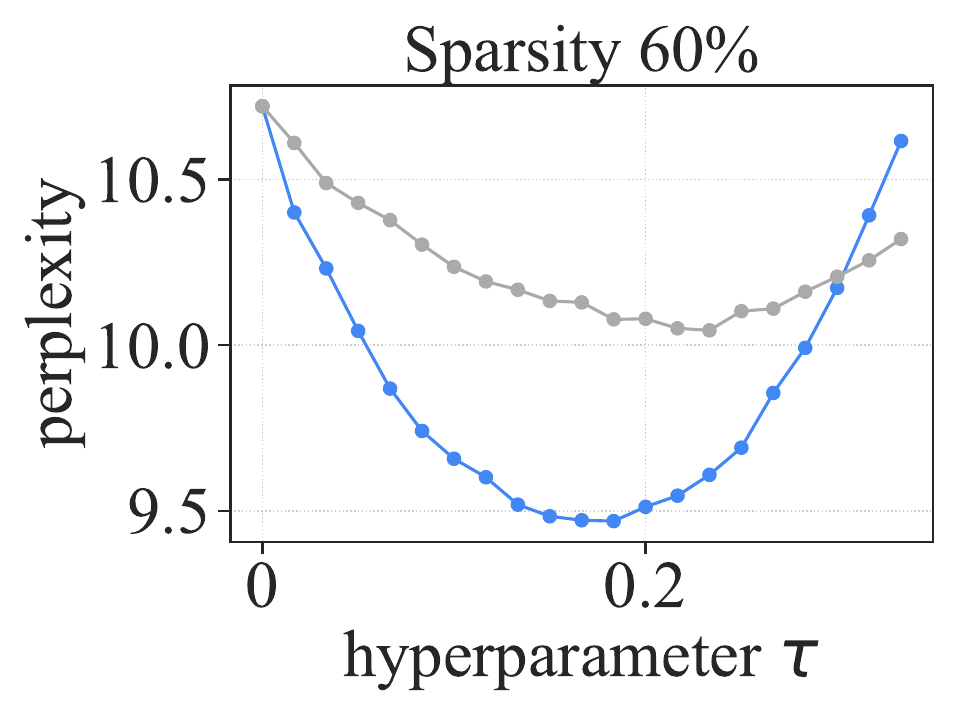}
    \includegraphics[width=0.24\linewidth,keepaspectratio]{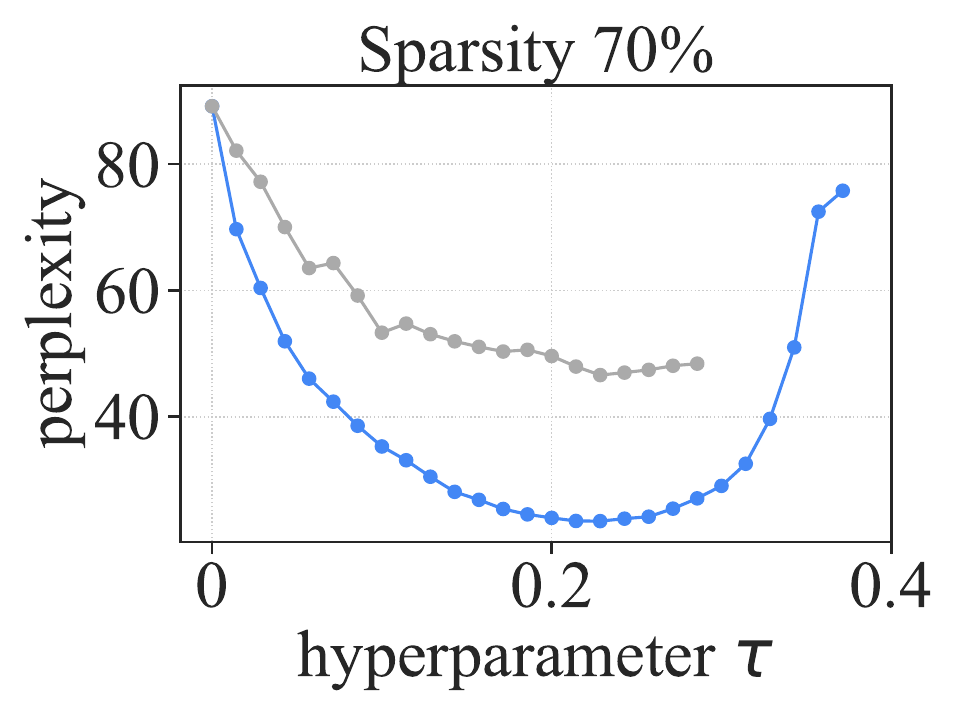}
    \includegraphics[width=0.24\linewidth,keepaspectratio]{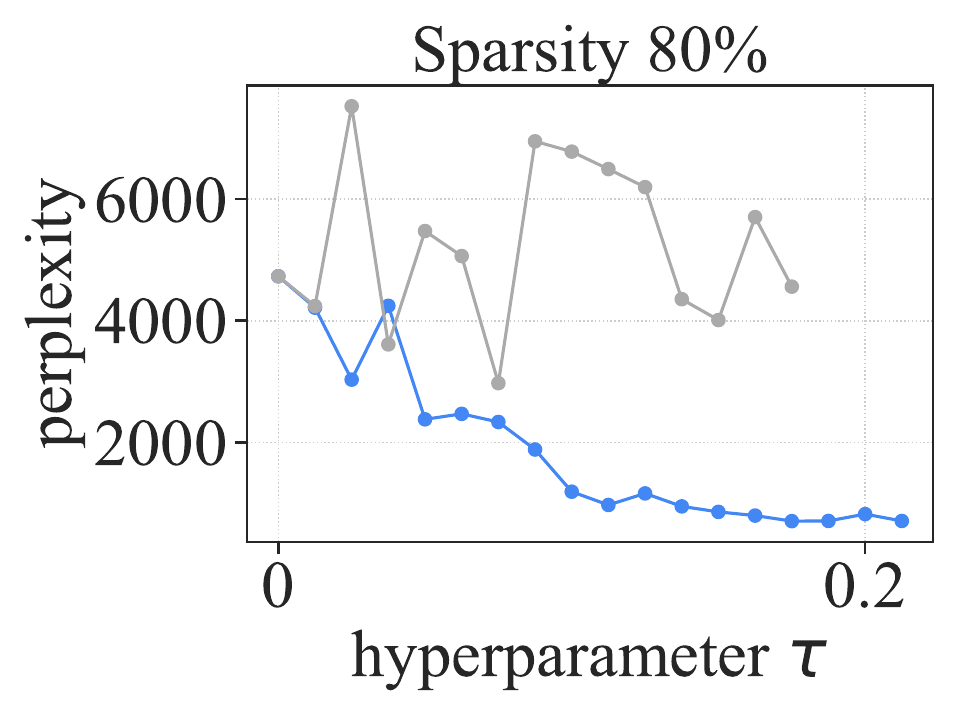}
    \caption{\textbf{Ablation study on sparsity allocation function hyperparameter $\tau$.} We use Wanda to prune LLaMA model using \ourmethod with both per-matrix and per-block allocation methods.} ~\label{fig:ablation_hyper}
\end{figure}

\subsection{Per-matrix vs. per-block}\label{sec:app-per-matrix-block}
In Figure~\ref{fig:ablation_hyper}, we compared per-matrix and per-block sparsity allocation across different values of the hyperparameter $\tau$, using Wanda.
The results show a regime transition in pruning effectiveness between the two methods. 
For sparsity levels below 50\%, the per-matrix approach (gray lines) achieves lower perplexity, indicating better performance. 
However, for sparsity values of 50\% and above, the per-block method (blue line) performs better.
We further focused on high sparsity 70\%, as higher sparsity is more important to provide efficiency improvements. Specifically, we evaluated both per-matrix and per-block methods in combination with various intra-layer pruning techniques.
Table~\ref{tb:layer-vs-block} shows that the per-block method consistently outperforms the per-matrix method when used in conjunction with three intra-layer pruning techniques.

Additionally, we analyzed the differences between MLP and Attention matrices by visualizing average \ALPHAHILL values for seven types of weight matrices in LLaMA-7B, as shown in Figure~\ref{fig:type-wise-alpha}.
The results indicate that query and key matrices have lower \ALPHAHILL values, suggesting they are less prunable. 
Based on this insight, we developed a new allocation strategy called ``Mixed'', which combines per-block and per-matrix approaches. 
As defined, Mixed first assigns a block-wise pruning ratio using the per-block method, then refines it within each block using per-matrix allocation. 
Table~\ref{tb:layer-vs-block} demonstrates that this Mixed approach provides further marginal improvements over both per-matrix and per-block methods.

\begin{minipage}{\textwidth}
\begin{minipage}[t]{0.52\textwidth}
\makeatletter\def\@captype{table}
\vspace{13mm}
    \resizebox{1.0\textwidth}{!}{
    \begin{tabular}{lcccc}
        \toprule
        Method & Uniform & Per-matrix & Per-block & Mixed \\
        \midrule 
        Magnitude & 48419.13 &	7384.13 & 231.01 & \bf{177.86} \\
        Wanda & 85.77 & 58.71 & 23.86 & \bf{22.46} \\
        SparseGPT & 26.30 & 25.10 & 18.54 & \bf{18.32} \\
        \bottomrule
    \end{tabular}
    }
    \caption{Comparing perplexity of sparse LLaMA-7B (sparsity=70\%) pruned by four types of sparsity allocation method.
    }\label{tb:layer-vs-block}
\vspace{-5mm}
\end{minipage}
\hspace{0.03\textwidth}
\begin{minipage}[t]{0.45\textwidth}
\vspace{2mm}
\makeatletter\def\@captype{figure}
    \centering
    \includegraphics[width=0.8\linewidth,keepaspectratio]{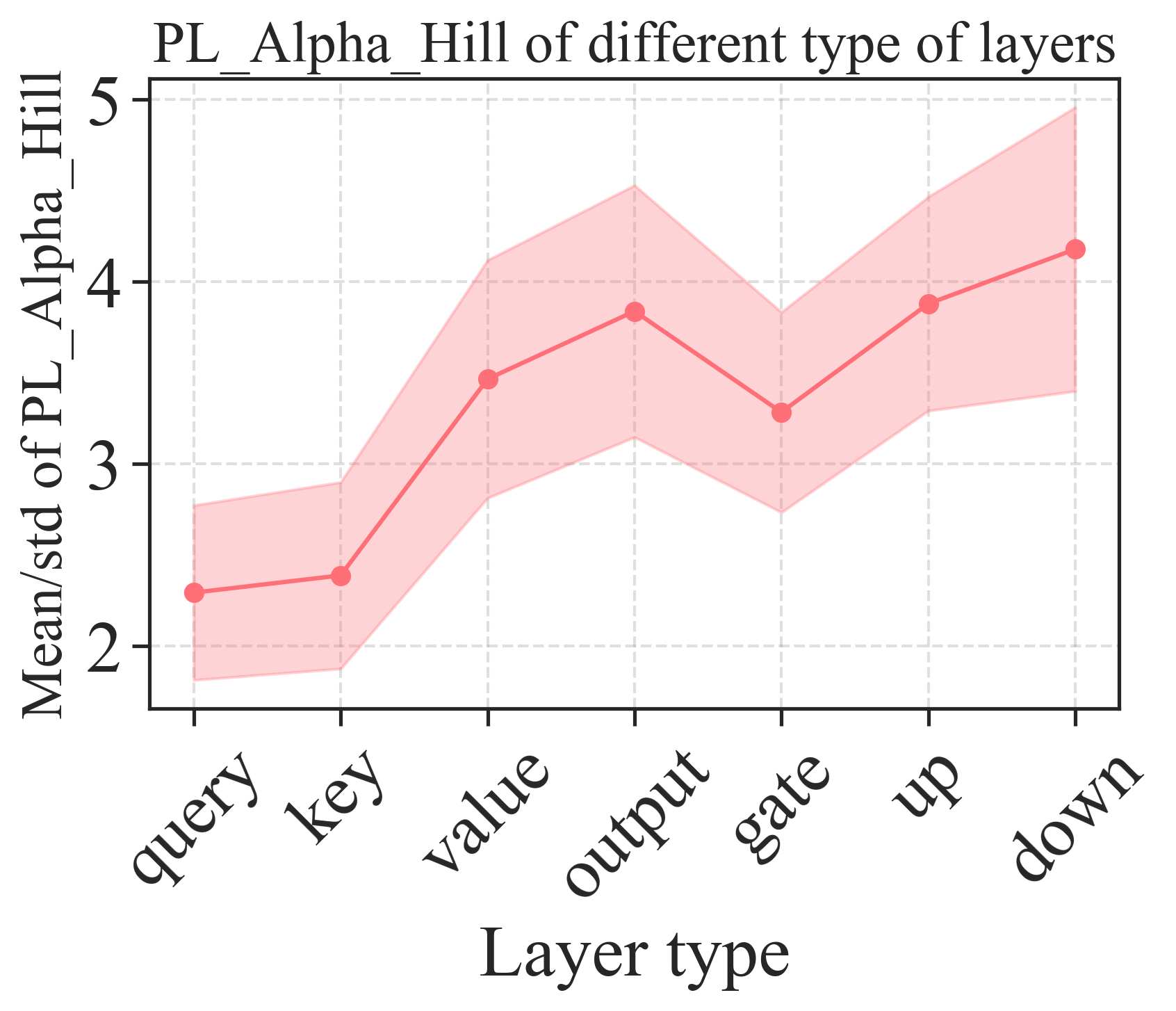} 
    \caption{Averaged \ALPHAHILL metric values for seven types of weight matrices in LLaMA-7B.
    }
    \label{fig:type-wise-alpha}
\vspace{-5mm}
\end{minipage}
\end{minipage}

\subsection{Different mapping functions}\label{app:diff-map-func}
\begin{figure}[!thb]
    \centering
    \begin{subfigure}{0.35\linewidth} 
    \includegraphics[width=\linewidth,keepaspectratio]{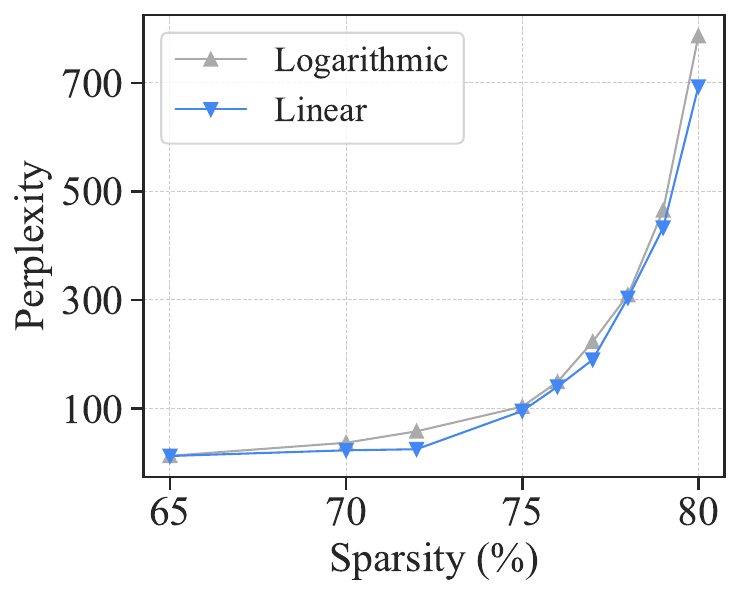} 
    \caption{Wanda}
    \end{subfigure}    
    \begin{subfigure}{0.35\linewidth} 
    \includegraphics[width=\linewidth,keepaspectratio]{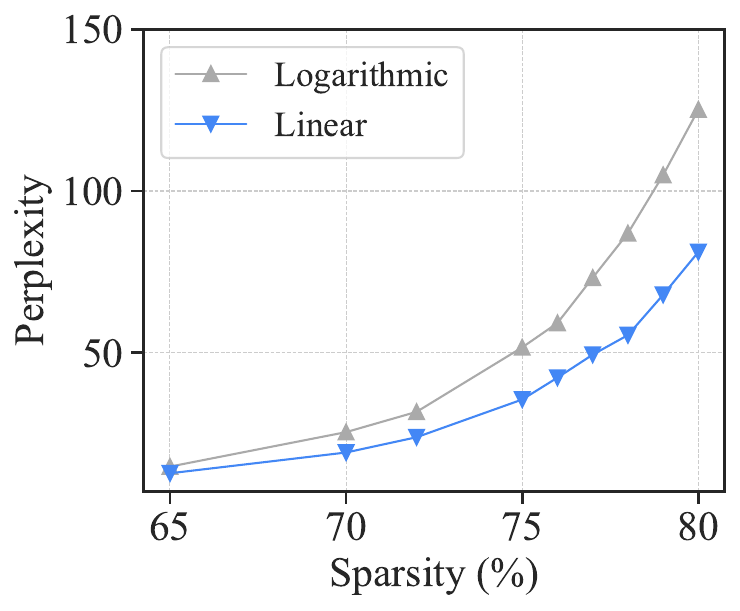} 
    \caption{SparseGPT}
    \end{subfigure}
\caption{\textbf{Comparing different mapping functions.} Linear (ours) refers to the linear mapping function that is used in our current method. Logarithmic refers to computing the logarithmic of metrics before linear mapping.  The model is LLaMA-V1-7B. (a-b) refers to combining different intra-layer pruning methods.} ~\label{fig:mapping-function}
\end{figure}

We provide ablation studies on the choices of sparsity assignment function. We implemented the logarithmic method and compared it with the linear mapping function used in our current approach, as shown in Figure~\ref{fig:mapping-function}. The results show that both methods perform similarly when combined with Wanda, but linear mapping slightly outperforms the proposed logarithmic method when combined with SparseGPT.

\FloatBarrier
\section{Hypterparamter setting}\label{sec:supp-hyper}
Here, we provide the values of $\tau$ used in the experiments, as shown in Table~\ref{tb:hypers}.

\begin{table}[!thb]
\centering
\begin{subtable}{0.45\textwidth}
    \centering
    \resizebox{1.05\textwidth}{!}{
    \begin{tabular}{lccc}
      \toprule
      & \multicolumn{3}{c}{Method} \\
      \cmidrule{2-4} 
      \multirow{-2}{*}{Model} & Magnitude & Wanda & SparseGPT \\
      \midrule
      LLaMA-7B & 0.3 & 0.2 & 0.2  \\
      LLaMA-13B & 0.5 & 0.3 & 0.3 \\
      LLaMA-30B & 0.4 & 0.3 & 0.3 \\
      LLaMA-65B & 0.4 & 0.3 & 0.3 \\
      \midrule
      LLaMA2-7B & 0.5 & 0.3 & 0.2 \\
      LLaMA2-13B & 0.4 & 0.3 & 0.3 \\
      LLaMA2-70B & 0.4 & 0.3 & 0.3 \\
      \bottomrule
\end{tabular} \label{tb:hyper-1}
}
\end{subtable}
\hspace{0.03\textwidth}
\begin{subtable}{0.45\textwidth}
    \centering
    \resizebox{1.0\textwidth}{!}{
    \begin{tabular}{lccccc}
      \toprule
      & \multicolumn{5}{c}{Sparsity} \\
      \cmidrule{2-6}
      \multirow{-2}{*}{Model} & 40\% & 50\% & 60\% & 70\% & 80\% \\
      \midrule
      ViT-B & 0.3 & 0.3 & 0.3 & -- & -- \\
      ViT-L & 0.3 & 0.3 & 0.3 & -- & -- \\

      \midrule
      DeiT-S & 0.4 & 0.4 & 0.4 & -- & -- \\
      DeiT-L & 0.3 & 0.3 & 0.3 & -- & -- \\

      \midrule
      ConvNext & 0.1 & 0.1 & 0.2 & 0.2 & 0.2 \\
      
      \bottomrule
    \end{tabular}
    \label{tb:hyper-2}
    }
\end{subtable}
\vspace{0.03\textwidth}
\caption{{\bf Left:} Hyperparameters setting for results in Section~\ref{sec:main-compare-baseline}. We report the optimal $\lambda$ after a small hyperparameter sweep within the range of $\tau \in$ [0.2, 0.3, 0.4, 0.5]. 
{\bf Right:} Hyperparameters setting for results in Section~\ref{sec:main-corro} for Vision Transformers at 40\%, 50\%, 60\% sparsity. We report the optimal $\lambda$ after a small hyperparameter sweep within the range of $\tau \in$ [0.1, 0.2, 0.3, 0.4, 0.5].}
\label{tb:hypers}

\end{table}

\FloatBarrier

\section{More baseline comparison}\label{sec:supp-baselines}
For assigning layerwise sparsity ratios, we compare \ourmethod with other methods. In this section, we provide definitions and details of these methods:
\begin{itemize}[noitemsep,topsep=0pt,leftmargin=*,after=,before=]
    \item 
    {\bf Uniform.~\cite{zhu2017prune}} 
    Every layer pruned with the same target sparsity.
    \item 
    {\bf Global~\cite{frankle2018lottery}.} 
    A global threshold uniformly applied to all layers to satisfy the overall sparsity requirement.
    The specific layerwise sparsity is automatically adjusted based on this threshold.
    \item 
    {\bf ER~\cite{mocanu2018scalable}.} 
    The sparsity of the convolutional layer is scaled proportionally to $1 - \frac{h^{l-1} + h^{l}}{h^{l-1} \times h^{l}}$ where $h^l$ refers to the number of neurons/channels in layer $l$.
    \item 
    {\bf ER-Plus~\cite{liu2022unreasonable}.} 
    ER-Plus modifies ER by forcing the last layer as dense if it is not, while keeping the overall parameter count the same.
    \item 
    {\bf LAMP~\cite{lee2020layer}.}
    This method modifies Magnitude-based pruning by rescaling the importance scores in a layer by a factor dependent on the magnitude of surviving connections in that layer. 
    \item
    {\bf OWL~\cite{yin2023outlier}.} 
    A non-uniform layerwise sparsity based on the distribution of outlier activations in LLMs, probing the possibility of pruning LLMs to high sparsity levels.
\end{itemize}
We adopt Wanda and SparseGPT as the pruning approach. The results are presented in Table~\ref{tb:layerwise-baseline-wanda} and Table~\ref{tb:layerwise-baseline-sparsegpt}, which indicate that \ourmethod significantly outperforms all baseline methods in relatively high-sparsity regimes.
Besides, we have conducted additional experiments using the ``layerwise error thresholding'' method, where each layer is pruned sequentially as specified in ~\cite{zhuang2018discrimination, ye2020good}. We have also implemented the rank selection method as specified in Section 5.2 of ~\cite{el2022data}. We present the updated experimental results in Table~\ref{tb:more-baselines}. We observe that our method outperforms all the other baselines.

\begin{table*}[!thb]
    \centering
    \resizebox{0.8\textwidth}{!}{
    \begin{tabular}{cccccccccc}
        \toprule
        
        Method/Perplexity ($\downarrow$) & 10\% & 20\% & 30\% & 40\% & 50\% & 60\% & 70\% & 80\% \\
        \midrule 
        Global & 14.11 & 3134 & 10293 & 10762 & 14848 & 17765 & 5147 & 39918.56 \\
        LAMP & 5.69 & 5.78 & 5.98 & 6.3912 & 7.57 & 12.86 & 185.52 & 15647.87 \\ 
        LAMP (per-block) & 5.70 & 5.82 & 6.00 & 6.40 & 7.25 & 10.95 & 98.77 & 7318.08 \\ 
        ER & 5.70 & 5.81 & 6.03 & 6.57 & 7.80 & 12.41 & 119.66 & 6263.79 \\
        ER-Plus & 5.70 & 5.82 & 6.05 & 6.62 & 8.00 & 14.04 & 229.17 & 6013.91 \\
        Uniform & 5.70 & 5.82 & 6.00 & 6.39 & 7.26 & 10.63 & 84.52 & 5889.13\\
        OWL & 5.70 & 5.80 & 6.01 & 6.39 & 7.22 & 9.35 & 24.56 & 1002.87 \\
        Ours & 5.69 & 5.81 & 6.00 & 6.37 & 7.18 & 9.47 & \bf{23.86} & \bf{698.56} \\
        \bottomrule
    \end{tabular}
    }
    \caption{WikiText validation perplexity ($\downarrow$) of  LLaMA-7B pruned by different allocation methods at various global sparsities using Wanda. 
    \ourmethod outperforms other layerwise sparsity at high sparsity range. 
    } \label{tb:layerwise-baseline-wanda}
\end{table*}

\begin{table*}[!thb]
    \centering
    \resizebox{0.8\textwidth}{!}{
    \begin{tabular}{ccccccccc}
        \toprule
        Method/Perplexity ($\downarrow$)  & 10\% & 20\% & 30\% & 40\% & 50\% & 60\% & 70\% & 80\% \\
        \midrule 
        LAMP & 5.69 & 5.78 & 5.96 & 6.34 & 7.37 & 11.27 & 31.96 & 274.73 \\ 
        LAMP (per-block) & 5.70 & 5.80 & 5.96 & 6.33 & 7.22 & 10.45 & 27.05 & 224.32 \\ 
        ER & 5.70 & 5.81 & 6.02 & 6.49 & 7.54 & 11.29 & 30.20 & 258.63 \\
        Uniform & 5.69 & 5.89 & 5.96 & 6.32 & 7.22 & 10.56 & 26.30 & 188.11 \\
        OWL & 5.71 & 5.81 & 5.97 & 6.35 & 7.22 & 9.51 & 19.49 & 84.94 \\
        Ours & 5.69 & 5.81 & 5.99 & 6.36 & 7.30 & 9.73 & \bf{18.44} & \bf{81.98} \\
        \bottomrule
    \end{tabular}
    }
    \caption{ Perplexity ($\downarrow$) of pruning LLaMA-7B into various global sparsities using SparseGPT. We compare our method with three other baseline sparsity allocation methods.
    } \label{tb:layerwise-baseline-sparsegpt}
\end{table*}

\begin{table*}[!thb]
    \centering
    \resizebox{0.8\textwidth}{!}{
    \begin{tabular}{cccc}
        \toprule
        Method & Global sparsity & Perplexity ($\downarrow$) & Zero-shot Accuracy ($\uparrow$) \\
        \midrule 
        Uniform                       & 70\% & 26.30 & 41.52 \\
        Layerwise error thresholding  & 70\% & 32.54 & 41.24 \\
        Rank selection                & 70\% & 21.34 & 42.90 \\
        Ours                          & 70\% & \bf{18.54} & \bf{45.48} \\
        \bottomrule
    \end{tabular}
    }
    \caption{Comparing our method to other layerwise sparsity baseline methods in pruning LLaMA-7B into 70\% sparsity. The perplexity is evaluated on the WikiText validation set. The zero-shot accuracy is averaged over 7 downstream tasks. Each method is combined with SparseGPT.
    } \label{tb:more-baselines}
\end{table*}

\FloatBarrier

\section{Complementary Experimental Results}\label{sec:supp-results}

\subsection{Shape metrics versus scale metrics on Vision Transformers}\label{sec:supp-vit}
We further evaluate different metrics for computing layerwise sparsity on Vision Transformers. Shape metrics, including \ALPHAHAT, \ENTROPY, \ALPHAHILL, and \STABLERANK, are obtained from the shapes of the ESDs. Scale metrics, including \NORM and \SPECTRALNORM, are norm-based metrics measuring the scale of weights matrices (which can also be obtained from the ESD). The results shown in Table~\ref{tb:shape-scale-cv} align with the results in LLMs that shape metrics outperform scale metrics on allocating layerwise sparsity and \ALPHAHILL performs the best.

\begin{table*}[!thb]
    \centering
    \begin{tabular}{c|ccc|ccc}
        \toprule
        Metric used for &\multicolumn{3}{c|}{ViT-L 16/224 accuracy ($\uparrow$)} & \multicolumn{3}{c}{DeiT-S 16/224 accuracy ($\uparrow$)} \\
        layerwise pruning ratios & 40\% & 50\% & 60\% & 40\% & 50\% & 60\% \\
        \midrule  
        Uniform  & 76.05 & 68.73 & 39.45 & 75.62 & 68.98 & 50.49  \\
        \midrule  
        \cellcolor{smallColorT} \NORM &  76.15 & 68.93 & 37.46  & 76.62 & 72.21 & 58.93 \\
        \cellcolor{smallColorT} \SPECTRALNORM  & 75.99 & 67.72 & 32.98 & 75.97 & 70.13 & 53.76   \\
        \midrule 
        \cellcolor{largeColorT} \ENTROPY & 76.37 & 71.17 & 49.09 & 76.77 & 71.63 & 58.23 \\
        \cellcolor{largeColorT} \STABLERANK & 75.91 & 68.93 & 41.26 & 75.63 & 68.64 & 46.86 \\
        \cellcolor{largeColorT} \ALPHAHAT & \bf{77.11} & 72.01 & 52.56 & 76.94 & 72.10 & 59.78 \\
        \cellcolor{largeColorT} \ALPHAHILL & 76.86 & \bf{72.12}  & \bf{55.62} & \bf{77.07} & \bf{72.38} & \bf{60.92} \\
        \bottomrule
    \end{tabular}
    \caption{{\bf Evaluating shape metrics versus scale metrics on allocating layerwise sparsities on Vision Transformers.} \colorbox{largeColorT}{\textit{Shape metrics}} are obtained from the shapes of the ESDs. \colorbox{smallColorT}{\textit{Scale metrics}} are norm-based metrics measuring the scale of weights matrices (which can also be obtained from the ESD). We choose two models, ViT-L and DeiT-S, and the results are shown on ImageNet-1K accuracy without fine-tuning. We observe that shape metrics outperform scale metrics and \ALPHAHILL performs the best.} \label{tb:shape-scale-cv}
\end{table*}

\subsection{More results on other efficiency metrics}\label{sec:app-other-eff}
\begin{figure*}[!thb]
    \centering
    \begin{subfigure}{0.37\linewidth}
    \includegraphics[width=\linewidth,keepaspectratio]{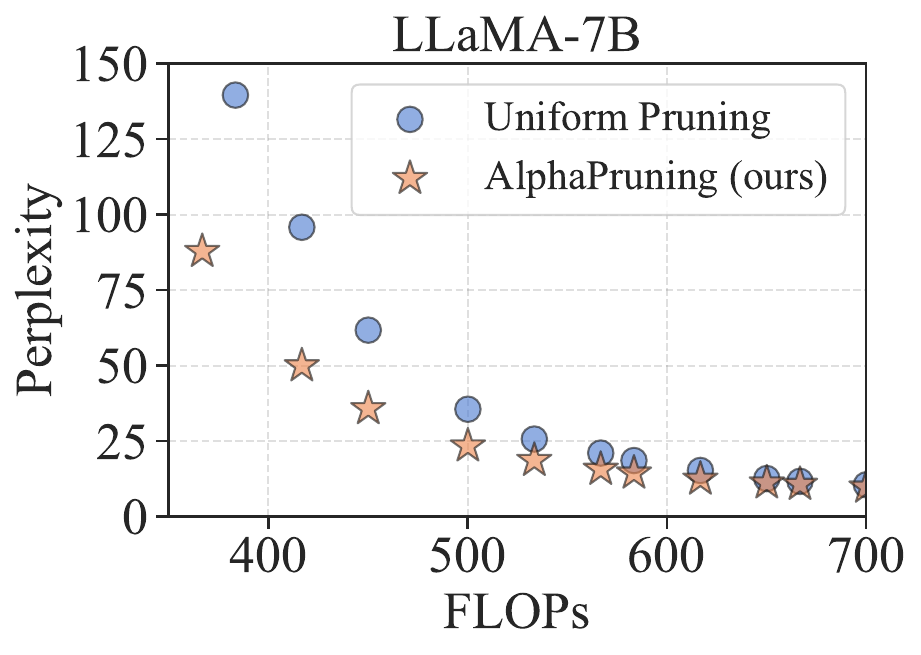}
    \caption{Broader perplexity range}
    \end{subfigure} 
    \centering 
    \begin{subfigure}{0.55\linewidth}
    \includegraphics[width=\linewidth,keepaspectratio]{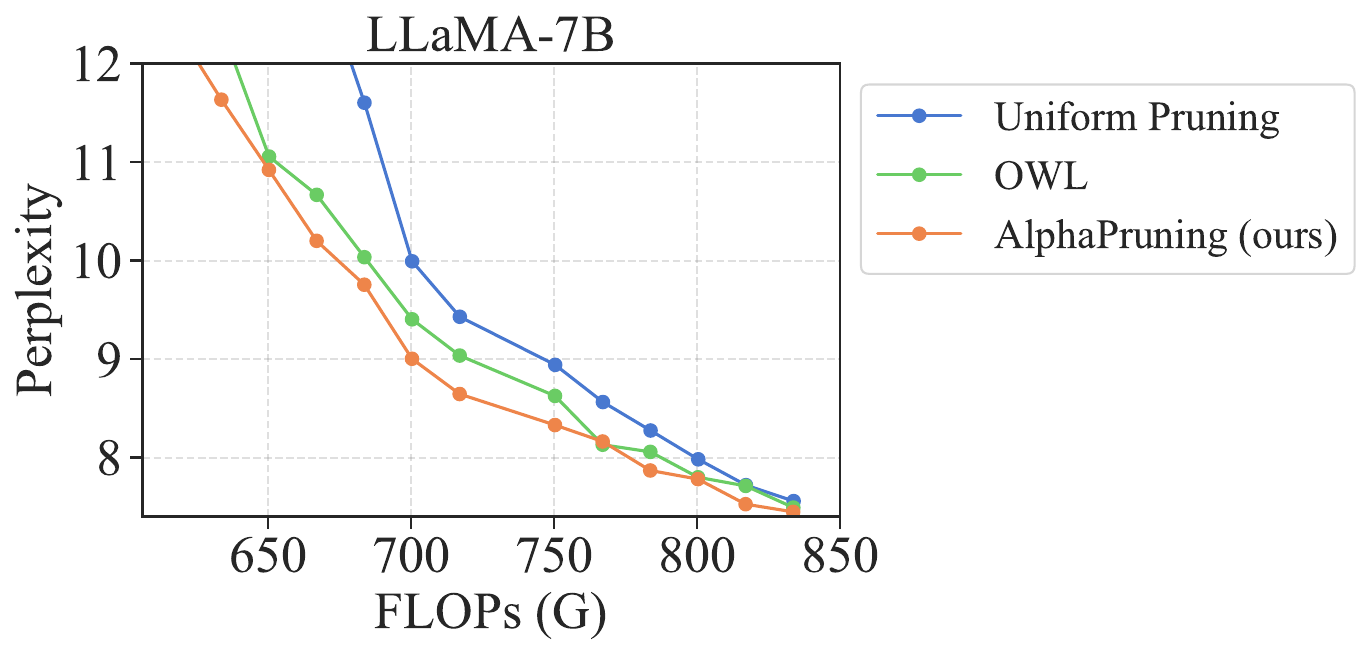} 
    \caption{Lower perplexity range}
    \end{subfigure} 
    \caption{Additional results of FLOPs measurement on the LLaMA-7B pruned by Uniform and our method with SparseGPT.
    }
    \label{fig:flops}
\end{figure*}

To further demonstrate the benefits of our approach, we provide results in other practical efficiency metrics such as FLOPs. Compared with uniform sparsity ratios, our approach is able to achieve a better performance-FLOPs trade-off. We have provided new results of FLOPs in Figure~\ref{fig:flops}.

Table~\ref{tb:flops} summarizes the results from Figure~\ref{fig:flops}. We show that, compared to uniform pruning, our method can achieve significant FLOPs reduction when pruned LLMs are compared at similar perplexity. 

\begin{table}[!h]
    \centering
    \resizebox{0.8\linewidth}{!}{
    \begin{tabular}{ccccccccc}
        \toprule
        Uniform  & Perplexity ($\downarrow$) & 9.80 & 18.58 & 21.06 & 25.70 & 35.57 & 61.76 & 95.82 \\
                & FLOPs ($\downarrow$) & 716.80 & 583.49 & 566.82 & 533.49 & 500.16 & 450.17 & 416.84 \\  
        \midrule
        Ours & Perplexity ($\downarrow$) & 9.78 & 15.67 & 18.63 & 23.39 & 35.74 & 49.94 & 87.81 \\
             & FLOPs ($\downarrow$) & 700.14 & 566.82 & 533.49 & 500.16 & 450.17 & 416.84 & 366.85 \\
        \bottomrule
    \end{tabular}}
    \vspace{3mm}
    \caption{Additional results of FLOPs measurement on the LLaMA-7B pruned by Uniform and our method with SparseGPT.} \label{tb:flops}
\end{table}

\subsection{Controlling the minimum layer sparsity for memory-limited hardware}
\label{sec:app-mini-spar}

To enhance the adaptability of our method to hardware, we demonstrate that our method allows for controlling the minimum sparsity, by adjusting the values of $s_1$ and $s_2$. 
Additionally, Table~\ref{tb:minimum_sparsity} demonstrates that increasing the minimum sparsity by changing $s_1$ and $s_2$  doesn't diminish the advantages of our method, compared to the uniform pruning baseline.

Recall that the sparsity of the layer of the model can be determined by Eqn.~\ref{eqn:sparsity-all}. Increasing $s_1$ leads to a higher minimum sparsity $\eta s_1$, while maintaining the same global sparsity $S$. 
Table \ref{tb:minimum_sparsity} displays the results of adjusting $s_1$ and $s_2$ to increase the minimum sparsity while maintaining the global sparsity at 70\%. 
At a minimum sparsity of 57\%, our method achieves the lowest perplexity. Even when the minimum sparsity is raised to 67\%, nearing the uniform pruning baseline, our method still attains a perplexity of 49.6, which is 36.17 points lower than that of uniform pruning (85.77).

\begin{table*}[!thb]
    \centering
    \resizebox{0.6\textwidth}{!}{
    \begin{tabular}{cccc}
        \toprule
        Method & Global sparsity $S$ & Minimum sparsity $\eta s_1$ & Perplexity \\
        \midrule 
        Uniform & 70\% & 70\% & 85.77 \\
        Ours & 70\%	& 50\% & 33.46 \\
        Ours & 70\%	& 55\% & 26.19 \\
        Ours & 70\%	& 57\% & \textbf{23.79} \\
        Ours & 70\%	& 60\% & 27.18 \\
        Ours & 70\%	& 65\% & 40.22 \\
        Ours & 70\%	& 67\% & \textbf{49.60} \\
        \bottomrule
    \end{tabular}
    }
    \caption{ Increasing the minimum sparsity of our method, while maintaining a global sparsity of 70\%, still yields performance improvements compared to a uniform sparsity ratio. We present the WikiText validation perplexity for LLaMA-7B pruned by both the Uniform method and our method, in conjunction with Wanda.
    } \label{tb:minimum_sparsity}
\end{table*}

\FloatBarrier
\subsection{Computational complexity of \ourmethod}\label{sup:comp-complexity}
The computational complexity of \ourmethod is not large because the most computation-intensive aspect of our method involves performing SVD decomposition on weight matrices, which can be further optimized through parallel processing. Table~\ref{tb:runtime} presents the runtime of our \ourmethod and an optimized version that uses parallel processing. The increase in runtime is reasonable, at 32.44\% with Wanda and 8.2\% with SparseGPT. These experiments were conducted on pruning LLaMA-7B to 70\% sparsity. The testing platform used A40 GPUs and an AMD EPYC 7713 64-Core CPU.

\begin{table*}[!thb]
    \centering
    \resizebox{0.9\textwidth}{!}{
    \begin{tabular}{cccc}
        \toprule
        Method & Ours & Original method (second) & Percentage of runtime increase (\%) \\
        \midrule 
        Wanda & -- & 176 & 0\% \\
        Wanda $w.$ Ours (1 GPU) & 462 & 178 & 259.55\% \\
        Wanda $w.$ Ours (8 GPUs) & 57 & 178 & 32.44\% \\
        \midrule 
        SparseGPT & -- & 699 & 0\% \\
        SparseGPT $w.$ Ours (1 GPU) & 462 & 695 & 66.47\% \\
        SparseGPT $w.$ Ours (8 GPUs) & 57 & 695  & 8.2\% \\
        \bottomrule
    \end{tabular}
    }
    \caption{ Runtimes of our method combined with Wanda and SparseGPT on LLaMA-7B.
    } \label{tb:runtime}
\end{table*}

\subsection{OPT family}
\label{other-llm-family}

In addition to LLaMA and LLaMA-2, we conduct experiments with OPT~\cite{zhang2023opt}.
Table~\ref{tb:opt-magnitude} shows the results of comparing our method to uniform sparsity when both are combined with magnitude pruning. Table~\ref{tb:opt-wanda-sparsegpt} shows the same comparison when both combined with Wanda~\cite{sun2023simple} /SparseGPT~\cite{frantar2023sparsegpt}. In most of the cases, our method outperforms the baseline method and achieves lower perplexity.

\begin{table*}[!thb]
    \centering
    \resizebox{0.8\textwidth}{!}{
    \begin{tabular}{cccccc}
        \toprule
        
        Method & Sparsity & OPT-125M & OPT-350M & OPT-2.7B & OPT-6.7B \\
        \midrule 
        magnitude & 40\% & 54.60 & 40.19 & 30.31 & 31.89 \\
        magnitude $w.$ Ours & 40\% & \bf{52.29} & \bf{38.20} & \bf{22.31} & \bf{19.53} \\
        magnitude & 50\% & 193.35 & 97.79 & 265.20 & 968.72 \\
        magnitude $w.$ Ours & 50\% & \bf{173.01} & \bf{95.61} & \bf{159.44} & \bf{224.91} \\
        \bottomrule
    \end{tabular}
    }
    \caption{The perplexity of OPT models pruned by uniform sparsity and our method combined with magnitude pruning. The perplexity is evaluated on WikiText validation set.
    } \label{tb:opt-magnitude}
\end{table*}

\begin{table*}[!thb]
    \centering
    \resizebox{0.8\textwidth}{!}{
    \begin{tabular}{cccccc}
        \toprule
        
        Method & Sparsity & OPT-125M & OPT-350M & OPT-2.7B & OPT-6.7B \\
        \midrule 
        Wanda & 70\% & 334.58 & 758.81 & 265.20 & 158.38 \\
        Wanda $w.$ Ours & 70\% & \bf{269.80} & \bf{654.17} & \bf{159.44} & \bf{40.81} \\
        SparseGPT & 70\% & 226.30 & 146.45 & 26.95 & 20.38 \\
        SparseGPT $w.$ Ours & 70\% & \bf{207.83} & \bf{136.90} & \bf{27.40} & \bf{20.31} \\
        \bottomrule
    \end{tabular}
    }
    \caption{The perplexity of OPT models pruned by uniform sparsity and our method combined with Wanda and SparseGPT. The perplexity is evaluated on WikiText validation set.
    } \label{tb:opt-wanda-sparsegpt}
\end{table*}

\subsection{More results on semi-structured and structured pruning} \label{sec:supp-structured}
To assess the potential of our non-uniform layerwise sparsity for hardware-friendly applications, we investigate \ourmethod across two distinct hardware-friendly pruning regimes: N:M sparsity and structured pruning. Following DominoSearch~\cite{sun2021dominosearch}, we study the mixed N:8 sparsity configuration. Instead of using a uniform N value across all layers, we allow individual layers to possess distinct N values while maintaining the same parameter count. We adopt \ourmethod to determine the optimal value of N for individual layers. The results shown in Table~\ref{tb:n-m} demonstrate that \ourmethod consistently outperforms the baselines.

\begin{table}[!thb]
    \centering
    \resizebox{0.75\linewidth}{!}{
    \begin{tabular}{llcccc}
        \toprule
        Method & Model &Layerwise Sparsity & 4:8 & 3:8 & 2:8 \\
        \midrule
        Wanda & LLaMA-7B & Uniform & 8.57 & 42.56 & 2962.00 \\
        Ours & LLaMA-7B & Mixed & \bf{8.55} & \bf{22.77} & \bf{585.81} \\
        \bottomrule
    \end{tabular}
    }
    \vspace{1mm}
    \caption{WikiText validation perplexity of pruned LLaMA-7B in Mixed N:8 sparsity configuration. The results are shown with Wanda and our non-uniform layerwise sparsity. Ours can lead to performance improvement at various sparsity levels.} \label{tb:n-m}
\end{table}

Furthermore, instead of pruning weights, we follow the recent methodology introduced in LLM Pruner~\cite{ma2023llm}, wherein entire neurons and attention heads are removed. This action facilitates the direct acceleration of pruned LLMs on GPUs or TPUs. We replace the uniform layerwise sparsity used by the LLM pruner with a non-uniform layerwise sparsity using \ourmethod. 
The results, shown in Table~\ref{tb:corro-LLMPruner}, demonstrate that \ourmethod can improve model performance at various sparsity levels.

\begin{table}[!h]
    \centering
    \resizebox{0.85\linewidth}{!}{
    \begin{tabular}{llccccc}
        \toprule
        Dataset & Pruning Method & Layerwise Sparsity & 20\% & 40\% & 60\% & 80\% \\
        \midrule
        WikiText & LLM Pruner & Uniform & 16.95 & 30.38 & 90.02 & 1228.17\\
        WikiText & LLM Pruner & Ours & \bf{16.78} & \bf{29.11}  & \bf{71.21} & \bf{952.77}  \\
        \midrule
        PTB & LLM Pruner & Uniform & 29.51 & 66.90 & 192.06 & 1691.87 \\
        PTB & LLM Pruner & Ours & \bf{29.11} & \bf{56.99} & \bf{144.97} & \bf{1002.40} \\
        \bottomrule
    \end{tabular}
    }
    \vspace{3mm}
    \caption{Applying \ourmethod to structured pruning method LLM-Pruner. The results are shown in WikiText validation perplexity of pruned LLaMA-7B at various sparsity levels.} \label{tb:corro-LLMPruner}
\end{table}

\begin{figure}[!thb]
    \centering
    \includegraphics[width=0.4\linewidth,keepaspectratio]{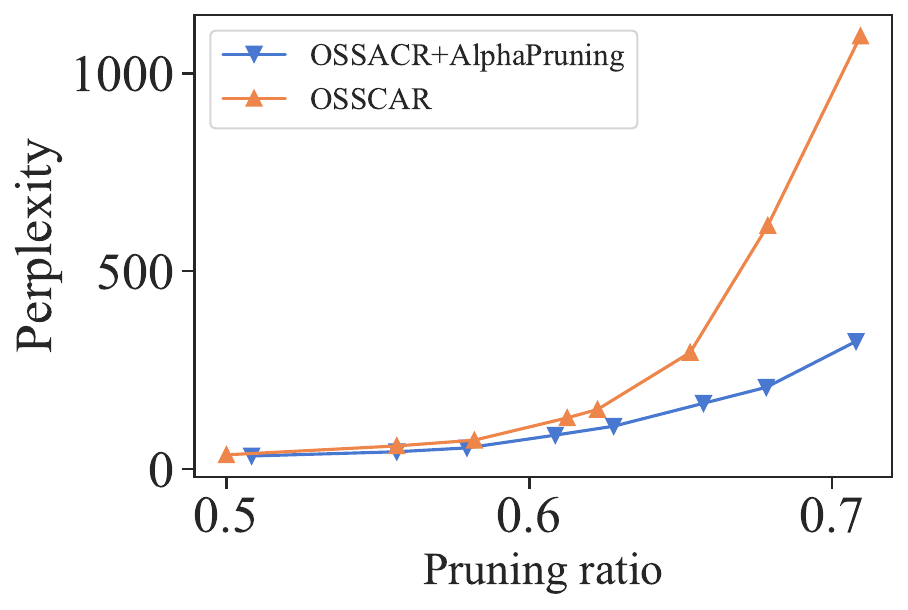} 
\caption{\textbf{Using \ourmethod to determine layerwise sparsity for OSSCAR.} The $x$-axis pruning ratio represents the fraction of pruned parameters relative to the total parameters in the linear sublayer of multi-head attention and the second sublayer of the feed-forward network, any other type of sublayers are not included. The model used is OPT-6.7B, and perplexity ($\downarrow$) is evaluated on WikiText.}~\label{fig:osscar}
\end{figure}

Another structured pruning method in LLM is OSSCAR~\cite{meng2024osscar}, which formulates the structured pruning problem as a quadratic program with combinatorial constraints. We integrated \ourmethod with OSSCAR, and we provide the results in Figure~\ref{fig:osscar}. OSSCAR prunes only the linear sublayer of multi-head attention and the second sublayer of the feed-forward network, applying uniform pruning across each transformer block. By incorporating \ourmethod's layer-wise sparsity allocation, we achieved non-uniform block-wise pruning ratios while keeping the global pruning ratio the same. The results show that integrating \ourmethod with OSSCAR can reduce perplexity at different sparsities. 

\subsection{Mixed-Precision Quantization}\label{sec:app-quantization}
We provide additional results to show that our method can enhance mixed-precision quantization by allocating precision to layers. We still use the \ALPHAHILL metric to estimate the heavy-tail extent of each layer, and more heavy-tailed layers are then allocated with higher precision.

Table~\ref{tb:quantization} shows that, compared to three baselines (random assignment, assigned by the norm of weights, and OWL), our \ourmethod method can always achieve the lowest perplexity under three types of mixed-precision quantization. Our experimental setup follows \citet{tang2022mixed}.

\begin{table}[!h]
    \centering
    \resizebox{0.65\linewidth}{!}{
    \begin{tabular}{cccc}
        \toprule
        Method & Mixed 3/4 Bit & Mixed 2/3/4 Bit & Mixed 2/4 Bit \\
        \midrule
        Random & 12.04 & 11455.54 & 14817.12 \\
        $L1$ norm & 14.61 & 13959.42 & 33679.21 \\
        OWL & 9.54 & 311.95 & 8429.39 \\
        Ours & \bf{9.01} & \bf{261.39} & \bf{7630.14} \\
        \bottomrule
    \end{tabular}
    }
    \vspace{3mm}
    \caption{Perplexity ($\downarrow$) of different methods on allocating precision to different layers for mixed-precision quantization with LLaMA-7B on WikiText.} \label{tb:quantization}
\end{table}

\FloatBarrier

\subsection{More fine-tuning results}\label{sec:app-fine-tuning}

Here, we provide results for LoRA fine-tuning pruned LLaMA-7B with a sparsity of 70\% using SparseGPT. We compare \ourmethod with Uniform and OWL. The experiment settings align with Section~\ref{sec:main-corro}. Table~\ref{tb:corro-lora-appen-1} summarizes the results for perplexity and mean zero-shot accuracies after fine-tuning pruned LLaMA-7B models, which show that the performance improvement achieved by our method doesn't diminish after fine-tuning.

\begin{table}[!h]
    \centering
    \resizebox{0.8\linewidth}{!}{
    \begin{tabular}{cccccc}
        \toprule
        Method & Sparsity & Fine-tuning & Perplexity ($\downarrow$) & Zero-shot accuracy ($\uparrow$) \\
        \midrule
        Dense model & -- & \usym{2717} & 5.68 & 60.08 \\
        \midrule
        Uniform & 70\% & \usym{2717} & 26.30 & 41.52 \\
        Uniform & 70\% & LoRA & 11.21 & 49.00 \\
        \midrule
        OWL & 70\% & \usym{2717} & 19.49 & 44.65  \\
        OWL & 70\% & LoRA & 11.13 & 49.33 \\
        \midrule
        Ours & 70\% & \usym{2717} & 18.54 & 45.48 \\
        Ours & 70\% & LoRA & \bf10.95 & 49.51  \\
        \bottomrule
    \end{tabular}
    }
    \vspace{3mm}
    \caption{WikiText validation perplexity and mean zero-shot tasks accuracy of SparseGPT pruned LLaMA-7B at 70\% sparsity after LoRA fine-tuning on 30,000 C4 dataset samples.} \label{tb:corro-lora-appen-1}
\end{table}

\subsection{More results on vision models}\label{sec:app-cv-results}
Here, we present more CV task results. We choose three widely used non-uniform layer-wise sparsity methods in the CV context including Global~\cite{frankle2018lottery}, ERK~\cite{mocanu2018scalable}, LAMP~\cite{lee2020layer}. Here, we use four ImageNet-1K pre-trained models (ViT-L, ViT-B, DeiT-B, DeiT-S), and we prune them to different sparsities.

\begin{table*}[!th]
    \centering
    \resizebox{1.0\textwidth}{!}{
    \begin{tabular}{l|ccc|ccc|ccc|ccc}
        \toprule
        & \multicolumn{3}{c|}{ViT-B 16/224} & \multicolumn{3}{c|}{ViT-L 16/224} & \multicolumn{3}{c|}{DeiT-S 16/224} & \multicolumn{3}{c}{DeiT-B 16/224}\\
        Method & 40\% & 50\% & 60\% & 40\% & 50\% & 60\% & 40\% & 50\% & 60\% & 40\% & 50\% & 60\% \\
        \midrule 
        Uniform & 70.87 & 59.46 & 29.97 & 76.05 & 68.73 & 39.45 & 75.62 & 68.98 & 50.49 & 80.08 & 76.37 & 61.72\\
        Global~\cite{frankle2018lottery} & 66.81 & 45.75 & 8.09 & 75.42 & 65.12 & 29.14 & 74.17 & 65.77 & 38.87 & 79.94 & 75.09 & 57.01\\
        ERK~\cite{mocanu2018scalable} & 70.89 & 60.49 & 33.15 & 76.26 & 69.51 & 40.57  & 75.65 & 69.80 & 52.95 & 80.05 & 76.22 & 63.49 \\
        LAMP~\citet{lee2020layer} & 69.45 & 57.51 & 26.99 & 75.71 & 67.29 & 30.80 & 75.51 & 69.46 & 50.79 & 80.19 & 76.35 & 63.32 \\
        Ours & \bf{71.58} & \bf{64.29} & \bf{44.21} & \bf{76.86} & \bf{72.12} & \bf{55.62} & \bf{77.07} & \bf{72.38} & \bf{60.92} & \bf{80.21} & \bf{77.11} & \bf{64.56} \\
        \bottomrule
    \end{tabular}
    }
    \caption{ImageNet-1K Accuracy ($\uparrow$) with various layerwise sparsity using Magnitude-based pruning, without finetuning. The results are shown at 40\%, 50\%, and 60\% sparsity on ViT-B, ViT-L, DeiT-S, and DeiT-B models. Higher accuracy is better.
    } \label{tb:cv} 
\end{table*}

\subsection{Zero-shot tasks performance}\label{sup:task-wise-performance}
For zero-shot results in Section~\ref{sec:main-compare-baseline}, the 7 evaluated zero-shot tasks are: BoolQ~\cite{clark2019boolq}, RTE~\cite{wang2018glue}, HellaSwag~\cite{zellers2019hellaswag}, WinoGrande~\cite{sakaguchi2021winogrande}, ARC Easy and Challenge~\cite{clark2018think} and OpenbookQA~\cite{mihaylov2018can}. We show the task-wise performance in Table~\ref{tb:llama-zeroshot} and Table~\ref{tb:llama2-zeroshot}.

\section{Experiments Compute Resources}\label{sec:com-res}
We conducted all our experiments using NVIDIA L40 (40GB) GPUs. Specifically, we used a single NVIDIA L40 GPU for pruning the 7B and 13B models, 4 GPUs for the 30B models, and 8 GPUs for the 65B models. For the LoRA fine-tuning, we operated under a constrained computational budget, employing 2x 40GB GPUs for the 7B models. Detailed information on the computational complexity of \ourmethod is provided in Appendix~\ref{sup:comp-complexity}.

\begin{table*}[!thb]
    \centering
    \resizebox{\linewidth}{!}{
    \begin{tabular}{cccccccccc}
        \toprule
        Model & Method & BoolQ & RTE & HellaSwag & WinoGrande & ARC-e & ARC-c & OBQA & Mean  \\
        \midrule
        & Dense & 75.02 & 66.79 & 56.94 & 70.00 & 75.29 & 41.89 & 34.60 & 60.08\\
        \cline{2-10}
         & Magnitude & 38.29 & 52.71 & 26.65 & 51.38 & 26.68 & 19.80 & 11.60 & 32.30\\
         & OWL $w$. Magnitude & 37.86 & 52.71 & 27.32 & \bf{52.57} & 28.96 & \bf{22.01} & 13.60 & 33.57 \\
        Llama-V1 & Ours $w$. Magnitude & \bf{40.31} & \bf{53.69} & \bf{30.87} & 51.22 & \bf{36.24} & 21.76 & \bf{16.20} & \bf{35.67} \\
        \cline{2-10}
        7B & Wanda &  56.45 & 55.96 & 28.94 & 51.38 & 33.32 & 18.26 & 13.80 & 36.73\\
        & OWL $w$. Wanda & 62.63 & 58.84 & 34.89 & 58.64 & \bf{46.97} & 24.32 & \bf{17.60} & 43.41 \\
            & Ours $w$. Wanda & \bf{66.12} & \bf{59.93} & \bf{36.26} & \bf{62.35} & 43.90 & \bf{25.17} & 17.20 & \bf{44.42} \\
        \cline{2-10}
         & SparseGPT  & 64.65 & 53.69 & 33.73 & 56.59 & 42.93 & 22.10 & 17.60 & 41.52 \\
         & OWL $w$. SparseGPT & 66.15 & 52.71 & 37.08 & 62.35 & \bf{48.40} & 26.02 & 19.80 & 44.65 \\
         & Ours $w.$ SparseGPT & \bf{68.26} & \bf{55.96} & \bf{37.81} & \bf{64.01} & 46.17 & \bf{27.39} & \bf{18.80} & \bf{45.48}  \\
        \midrule 
        & Dense & 77.86 & 70.75 & 59.91 & 72.61 & 77.40 & 46.42 & 33.20 & 62.59 \\
        \cline{2-10}
         & Magnitude & 52.97 & \bf{50.54} & 26.58 & \bf{50.75} & 28.45 & 20.56 & 14.80 & 34.95 \\
         & OWL $w$. Magnitude & 55.75 & 49.10 & 27.80 & 50.28 & 31.23 & 23.89 & 20.00 & 36.86 \\
        Llama-V1 & Ours $w$. Magnitude & \bf{61.28} & 46.93 & \bf{30.24} & 50.43 & \bf{31.23} & \bf{26.28} & \bf{21.20} & \bf{38.23} \\
        \cline{2-10}
        13B & Wanda & 61.90 & 52.71 & 30.50 & 53.12 & 40.91 & 17.58 & 15.60 & 38.90\\
            & OWL $w$. Wanda & 62.81 & 52.71 & 38.57 & 63.46 & 57.07 & 26.37 & 20.40 & 45.91 \\
            & Ours $w$. Wanda  & \bf{64.83} & \bf{52.71} & \bf{41.04} & \bf{64.96} & \bf{59.39} & \bf{28.24} & \bf{21.20} & \bf{47.48} \\
        \cline{2-10}
         & SparseGPT & 66.76 & 52.70 & 37.08 & 63.06 & 53.07 & 26.02 & 21.00 & 45.67  \\
         & OWL $w.$ Sparsegpt & 66.82 & 53.07 & 40.36 & 66.22 & \bf{57.37} & 28.67 & 20.80 & 47.61 \\
         & Ours $w.$ Sparsegpt & \bf{67.58} & \bf{54.51} & \bf{42.43} & \bf{67.80} & 55.47 & \bf{29.10} & \bf{22.00} & \bf{48.41}  \\
        \midrule 
        & Dense & 82.69 & 67.15 & 63.35 & 75.77 & 80.43 & 52.90 & 36.00 & 65.47 \\
        \cline{2-10}
         & Magnitude & 39.30 & 46.21 & 25.77 & 52.49 & 25.29 & 21.50 & 16.20 & 32.39 \\
         & OWL $w$. Magnitude & 39.93 & \bf{58.48} & 25.94 & 52.88 & 27.31 & 18.60 & 14.00 & 33.88 \\
        Llama-V1 & Ours $w$. Magnitude & \bf{62.02} & 47.29 & \bf{32.61} & \bf{57.14} & \bf{47.18} & \bf{27.47} & \bf{24.20} & \bf{42.56} \\
        \cline{2-10}
        30B & Wanda & \bf{66.09} & 56.68 & 43.96 & 67.09 & 65.28 & 31.83 & 26.60 & 51.07 \\
        & OWL $w$. Wanda  & 65.02 & 49.46 & 47.69 & 69.77 & 68.98 & 36.62 & 29.20 & 52.38 \\
            & Ours $w$. Wanda  & 63.82 & \bf{58.12} & \bf{49.40} & \bf{71.27} & \bf{69.40} & \bf{38.31} & \bf{31.00} & \bf{54.48} \\
        \cline{2-10}
         & SparseGPT &  68.13 & \bf{61.01} & 44.50 & 68.75 & 65.70 & 33.53 & 27.60 & 52.75 \\
         & OWL $w.$ SparseGPT & 67.95 & 55.60 & 46.96 & 72.22 & 67.13 & \bf{35.49} & 26.80 & 53.16 \\
         & Ours $w.$ SparseGPT &  \bf{68.96} & 57.04 & \bf{47.49} & \bf{72.30} & \bf{67.97} & 34.73 & \bf{30.00} & \bf{54.07} \\
        \midrule 
        & Dense & 84.60 & 69.68 & 65.40 & 77.51 & 80.95 & 52.82 & 38.40 & 67.05 \\
        \cline{2-10}
         & Magnitude & 51.85 & 54.51 & 38.55 & 56.67 & 57.35 & 29.95 & 26.40 & 45.04 \\
         & OWL $w$. Magnitude & 70.30 & 52.71 & 50.01 & \bf{67.88} & 65.40 & 35.75 & \bf{31.80} & 53.42 \\ 
        Llama-V1 & Ours $w$. Magnitude & \bf{70.65} & \bf{62.45} & \bf{51.35} & 66.61 & \bf{66.50} & \bf{37.54} & 31.40 & \bf{55.22} \\
        \cline{2-10}
        65B & Wanda & 76.45 & 56.68 & 47.50 & 69.61 & 70.55 & 35.92 & 27.60 & 54.90 \\
        & OWL $w$. Wanda  & 78.50 & 58.48 & 50.90 & 74.11 & 70.70 & 38.05 & \bf{30.60} & 57.34 \\
            & Ours $w$. Wanda  & \bf{81.85} & \bf{61.01} & \bf{52.55} & \bf{75.06} & \bf{71.60} & \bf{40.02} & 30.40 & \bf{58.93} \\
        \cline{2-10}
         & SparseGPT &  81.00 & 58.84 & 50.70 & 74.66 & 70.90 & 40.19 & 28.40 & 57.81 \\
         & OWL $w.$ SparseGPT & 81.30 & 67.15 & 51.25 & \bf{74.98} & 68.35 & 37.71 & 27.00 & 58.25 \\
         & Ours $w.$ SparseGPT & \bf{83.80} & \bf{71.48} & \bf{52.40} & 74.59 & \bf{69.30} & \bf{38.05} & \bf{38.05} & \bf{59.72}  \\
        \bottomrule
    \end{tabular}
    }
    \caption{Accuracies (\%) of LLaMA for 7 zero-shot tasks with unstructured 70\% sparsity. We compare \ourmethod with uniform pruning ratios and OWL using Magnitude-based pruning, Wanda and SparseGPT.}
    \label{tb:llama-zeroshot}
\end{table*}

\begin{table*}[h]
    \centering
    \resizebox{\linewidth}{!}{
    \begin{tabular}{cccccccccc}
        \toprule
        Model & Method & BoolQ & RTE & HellaSwag & WinoGrande & ARC-e & ARC-c & OBQA & Mean  \\
        \midrule
        & Dense & 62.81 & 59.71 & 77.73 & 69.14 & 76.30 & 43.43 & 31.40 & 59.71 \\
        \cline{2-10}
         & Magnitude & 37.95 & \bf{53.07} & 25.96 & 49.25 & 27.74 & 22.70 & 17.00 & 33.38 \\
         & OWL $w$. Magnitude & 38.75 & 52.35 & 27.38 & 48.38 & \bf{32.28} & 22.27 & 17.80 & 34.17 \\
        Llama-V2 & Ours $w$. Magnitude & \bf{57.43} & 51.99 & \bf{28.03} & \bf{51.14} & 31.99 & \bf{23.81} & \bf{20.60} & \bf{37.86} \\
        \cline{2-10}
        7B & Wanda & 50.64 & 52.71 & 27.80 & 50.04 & 30.64 & 18.60 & 12.20 & 34.66 \\
        & OWL $w$. Wanda & 62.11 & 52.71 & 31.83 & 55.96 & \bf{43.52} & 20.31 & 16.80 & 40.46 \\
            & Ours $w$. Wanda & \bf{62.23} & \bf{52.71} & \bf{34.56} & \bf{60.85} & 43.27 & \bf{22.27} & \bf{18.00} & \bf{41.98} \\
        \cline{2-10}
         & SparseGPT  & 65.38 & 53.43 & 33.55 & 57.62 & 44.52 & 22.91 & 16.40 & 41.84 \\
         & OWL $w.$ SparseGPT  & \bf{66.94} & 52.71 & 36.57 & \bf{63.06} & 49.33 & \bf{24.49} & \bf{21.60} & \bf{44.96} \\
         & Ours $w.$ SparseGPT  & 65.93 & \bf{54.15} & \bf{36.83} & 62.19 & \bf{49.62} & 24.23 & 19.80 & 44.68  \\
        \midrule 
        & Dense & 80.61 & 65.34 & 60.04 & 72.22 & 79.46 & 48.46 & 35.20 & 63.05\\
        \cline{2-10}
         & Magnitude & 38.72 & 52.70 & 27.56 & 49.33 & 31.27 & 20.73 & 14.60 & 33.56 \\
         & OWL $w$. Magnitude & 38.69 & 52.71 & 34.59 & 54.30 & 38.97 & 23.81 & 15.80 & 36.98 \\
        Llama-V2 & Ours $w$. Magnitude & \bf{70.55} & \bf{52.71} & \bf{37.13} & \bf{62.98} & \bf{42.51} & \bf{27.22} & \bf{18.60} & \bf{44.53} \\
        \cline{2-10}
        13B & Wanda &  62.39 & 52.71 & 28.98 & 51.14 & 35.10 & 18.00 & 11.80 & 37.16\\
        & OWL $w$. Wanda & \bf{63.46} & 52.71 & 36.31 & 60.46 & \bf{55.64} & 24.91 & \bf{21.80} & 45.04 \\
            & Ours $w$. Wanda & 62.57 & \bf{54.87} & \bf{40.28} & \bf{67.32} & 54.46 & \bf{29.35} & 21.60 & \bf{47.21} \\
        \cline{2-10}
         & SparseGPT & 66.64 & 52.71 & 36.26 & 59.91 & 54.00 & 25.85 & 20.80 & 45.17 \\
         & OWL $w.$ SparseGPT & 68.04 & 54.15 & 39.31 & 65.75 & \bf{57.70} & 27.82 & 22.80 & 47.94 \\
         & Ours $w.$ SparseGPT & \bf{68.13} & \bf{57.04} & \bf{41.26} & \bf{68.03} & 57.15 & \bf{29.18} & \bf{24.00} & \bf{49.26}  \\
        \midrule 
        & Dense & 83.50 & 67.87 & 66.00 & 77.98 & 82.60 & 54.27 & 37.20 & 67.06 \\
        \cline{2-10}
         & Magnitude & 39.15 & \bf{54.51} & 42.55 & 57.22 & 56.35 & 31.57 & 24.20 & 43.65 \\
         & OWL $w$. Magnitude & \bf{64.45} & 53.51 & 45.45 & \bf{68.06} & 61.00 & \bf{33.98} & 27.00 & 50.49 \\
        Llama-V2 & Ours $w$. Magnitude & 63.10 & 53.07 & \bf{45.65} & 67.09 & \bf{62.55} & 33.96 & \bf{29.60} & \bf{50.72} \\
        \cline{2-10}
        70B & Wanda & \bf{74.30} & 60.65 & 49.10 & 74.43 & 71.85 & \bf{38.74} & 28.00 & 56.72 \\
            & Ours $w$. Wanda  & 73.65 & \bf{63.18} & \bf{51.40} & \bf{74.74} & \bf{72.70} & 38.13 & \bf{29.60} & \bf{57.63} \\
        \cline{2-10}
         & SparseGPT & \bf{80.75} & 63.90 & \bf{52.30} & \bf{75.77} & 73.60 & \bf{41.64} & 29.80 & 59.68  \\
         & OWL $w.$ SparseGPT & 79.25 & 64.26 & 51.95 & 74.98 & 73.00 & 40.53 & \bf{30.40} & 59.20   \\
         & Ours $w.$ SparseGPT & 80.40 & \bf{70.04} & 51.90 & 75.06 & \bf{74.10} & 40.87 & 29.20 & \bf{60.23}  \\
        \bottomrule
    \end{tabular}
    }
    \caption{Accuracies (\%) of LLaMA-2 for 7 zero-shot tasks with unstructured 70\% sparsity. We compare \ourmethod with uniform pruning ratios and OWL using Magnitude-based pruning, Wanda, and SparseGPT.}
    \label{tb:llama2-zeroshot}
\end{table*}

\clearpage

\end{document}